\newcommand{\CLASSINPUTinnersidemargin}{54pt}

 \documentclass[letterpaper, 10 pt, conference]{styles/ieee/IEEEconf}  % Conference with changes

\IEEEoverridecommandlockouts                              % This command is only needed if
\IEEEsettopmargin{t}{72pt}

\usepackage{algorithm}
\usepackage{algpseudocode}

\usepackage{cite}
\usepackage{styles/cns_style}
\usepackage{styles/paper_vars}

\usepackage{pbalance} % thanks for this suggestion!!!
\usepackage{enumitem}

\usepackage[caption=false]{subfig}

\usepackage{booktabs}

\usepackage{orcidlink}

\newcommand{\equt}[1]{~(\ref{#1})}
\newcommand{\equtt}[1]{~(#1)}
\newcommand{\equst}[1]{~(#1)}
\newcommand{\figt}[1]{Fig.~\ref{#1}}

\newcommand{\tabt}[1]{Table~\ref{#1}}

\newcommand{\sect}[1]{Sec.~\ref{#1}}
\newcommand{\sectt}[1]{Sec.~#1}
\newcommand{\appt}[1]{App.~\ref{#1}}

\definecolor{addgreen}{RGB}{0,204,0}
\definecolor{addblue}{RGB}{0,0,204}
\definecolor{remred}{RGB}{204,0,0}
\definecolor{difforange}{RGB}{255,105,0}
\newcommand{\add}[1]{{\color{addblue} #1}}
\newcommand{\rem}[1]{{\color{remred} #1}}
\NewDocumentCommand{\diff}{O{}m}{
    \ifthenelse{\isempty{#1}}
    {{\color{addblue} #2}}
    {{\color{difforange} \sout{#1} #2}}
}
\usepackage[commandnameprefix=ifneeded]{changes}
\renewcommand{\rem}[1]{}
\renewcommand{\add}[1]{#1}
\RenewDocumentCommand{\diff}{O{}m}{#2}

\renewcommand{\placeholdertext}[1]{} % <-- to disable it
\begin{document}

% \title{Equivariant filter design for multi-GNSS sensors}
% \title{Equivariant Filtering: Designing EqFs for everyday UAV use on the example of multi-GNSS setups}
\title{Revisiting multi-GNSS Navigation for UAVs --\\ An Equivariant Filtering Approach}

\author{%
% either
% Alessandro Fornasier$^{1*}$, Martin Scheiber$^{1*}$, and Stephan Weiss$^{1}$% <-this % stops a space
% or
% Martin Scheiber$^{1*}$, Alessandro Fornasier$^{1*}$, Christian Brommer$^{1}$, and Stephan Weiss$^{1}$% <-this % stops a space
Martin Scheiber$^{1*}$\,\orcidlink{0000-0002-3415-7378}, %
Alessandro Fornasier$^{1*}$\,\orcidlink{0000-0002-1774-3236}, %
Christian Brommer$^{1}$\,\orcidlink{0000-0002-2801-2172}, and %
Stephan Weiss$^{1}$\,\orcidlink{0000-0001-6906-5409} % for final version
\thanks{\vspace{-0.6cm}}% for pre-print version
\thanks{This work has received funding from the European Union’s Horizon 2020 research and innovation programme under grant agreement 871260 (BugWright2), and was sponsored by the Army Research Office under Cooperative Agreement Number W911NF-21-2-0245. The views and conclusions contained in this document are those of the authors and should not be interpreted as representing the official policies, either expressed or implied, of the Army Research Office or the U.S. Government. The U.S. Government is authorized to reproduce and distribute reprints for Government purposes notwithstanding any copyright notation herein.
}%
\thanks{$^{1}$All authors are with Institute of Smart Systems Technologies, Control of Networked Systems Group,
        University of Klagenfurt, Universitätsstraße 65-67, 9020 Klagenfurt, Austria.
        Email: {\tt\scriptsize \{first.lastname\}@ieee.org}.}%
%\thanks{$^{2}$This author is with Aperture Robotics, LLC.
%        {\tt\scriptsize christian@aperture.us}}%
\thanks{$^{*}$ A. Fornasier and M. Scheiber contributed equally.}%
% \thanks{Digital Object Identifier (DOI): see top of this page.} %
\thanks{\textbf{Pre-print version, accepted to ICAR23 in Oct/2023,  DOI follows ASAP~\copyright IEEE.}}% for pre-print version
}

\maketitle

%%%%%%%%%%%%%%%%%%%%%%%%%%%%%%%%%%%%%%%%%%%%%%%%%%%%%%%%%%%%%%%%%%%%%%%%%%%%%%%%
\begin{abstract}
%Abstract: max 200 words
    % Recent advances in the field of state estimation have introduced and shown the advantages of equivariant filtering: better state-error definition, better convergence (rates), and independence of initial state estimates. Yet, equivariant filters (EqFs) are not widely explored for uncrewed aerial vehicle (UAV) estimation.
    %
    % With this work, we present an EqF formulation for fusing the most common sensors in outdoor robotics equivariantly: global navigation satellite system (GNSS) sensors and an inertial measurement unit (IMU). Using a semi-direct symmetry group and its properties for deriving the filter equations, we can show improved performance of the EqF compared to similar multiplicative extend Kalman filter (MEKF) approaches. Especially the orientation can be estimated more precisely due to the better system and error definition.
    %
    % Finally, we evaluate our framework on all outdoor runs of the \emph{INSANE Dataset} to show the practical usability of EqFs in real-world environments.
    %
    In this work, we explore the recent advances in equivariant filtering for inertial navigation systems to improve state estimation for uncrewed aerial vehicles (UAVs). Traditional state-of-the-art estimation methods, e.g., the multiplicative Kalman filter (MEKF), have some limitations concerning their consistency, errors in the initial state estimate, and convergence performance.
    Symmetry-based methods, such as the equivariant filter (EqF), offer significant advantages for these points by exploiting the mathematical properties of the system - its symmetry. These filters yield faster convergence rates and robustness to wrong initial state estimates through their error definition.
    To demonstrate the usability of EqFs, we focus on the sensor-fusion problem with the most common sensors in outdoor robotics: global navigation satellite system (GNSS) sensors and an inertial measurement unit (IMU). We provide an implementation of such an EqF leveraging the semi-direct product of the symmetry group to derive the filter equations. %leveraging the mathematical properties of a semi-direct symmetry group to derive the filter equations.
    To validate the practical usability of EqFs in real-world scenarios, we evaluate our method using data from all outdoor runs of the \emph{INSANE Dataset}.  Our results demonstrate the performance improvements of the EqF in real-world environments, highlighting its potential for enhancing state estimation for UAVs.
\end{abstract}

% \begin{IEEEkeywords}
% \end{IEEEkeywords}

%%%%%%%%%%%%%%%%%%%%%%%%%%%%%%%%%%%%%%%%%%%%%%%%%%%%%%%%%%%%%%%%%%%%%%%%%%%%%%%%
% \section*{Supplementary Material}

%%%%%%%%%%%%%%%%%%%%%%%%%%%%%%%%%%%%%%%%%%%%%%%%%%%%%%%%%%%%%%%%%%%%%%%%%%%%%%%%
\vspace{-0.6em} %for pre-print version
\section{Introduction}
\label{sec:intro}
\vspace{-0.4em} %for pre-print version
% \rem{This is a template version, will be filling out the draft in the coming days}

% \rem{\paragraph{Storyline} Talk about the usability of EqFs in real robotic
% applications, such as a multi-GNSS UAV (DJI, etc.). Improved performance to
% EKFs, and better results. Thus can use this filter for real robotic applications
% and ...}

% \rem{Note: after talk with @Stephan: Idea, write maths as definition as if EqF and symmetry is clear to everyone (same as EKF nowadays) and just define everything but no proof. Then show contributions on GNSS results!}

% EqF have two advantageous properties: They are guaranteed to converge, they are consistent. <-- this is the story for why one should use eqfs rather then ekfs even if results are similar

%% intro start:
% --- needs cleanup and rewrite, will do

Since the dawn of autonomous micro-\acp{uav}, filter-based state estimation has played a major role for observer design. The \ac{ekf} and its variants provide the possibilities to safely navigate \acp{uav} in most environments by fusing measurements of (multiple) sensors. %However, these approaches can produce larger state errors in certain scenarios due to their erroneous (linearized) system dynamics~\cite{barrau_non-linear_2015}.
However, these approaches can suffer from large state errors in certain scenarios due to their simplified uncertainty handling and state-dependent linerarization of the non-linear system dynamics~\cite{barrau_non-linear_2015}.
% Does anybody knwo what the first paper with EKF for UAVs was?

In recent years, the aerial vehicles community has seen the advent of filter-based solutions that exploit the symmetries of the underlying dynamical system to design more robust filter. In this regard, the \ac{eqf} has emerged as general filter design technique for systems on homogeneous space with symmetries~\cite{mahony_equivariant_2020-1, van_goor_equivariant_2020, van_goor_equivariant_2023, mahony_observer_2022, ng_equivariant_2020}.
\Acp{eqf} possess several advantages compared to other filter methods.
By exploiting the system's underlying symmetry and lifting the system onto the symmetry group, the \acl{eqf} framework provides a natural choice of global state error with improved linearized error dynamics.
Because of this, \acp{eqf} has a wider basin of attraction yielding high robustness to wrong initial conditions, and filter convergence regardless of their initial estimate~\cite{fornasier_equivariant_2022}. Moreover, \acp{eqf} are shown to be naturally consistent even in the presence of unobservable states, which is in stark contrast to \ac{ekf}-like approaches~\cite{barrau_invariant_2018, wu_invariant-ekf_2017, fornasier_vinseval_2021} and even in contrast to the ``imperfect \ac{iekf}''~\cite{barrau_invariant_2017} where some states, in particular the sensor biases, are ``tacked-on'' to the symmetry due to the approach's limitations~\cite{fornasier_overcoming_2022}.

Nowadays, \ac{uav} manufacturers rely on multi-sensor fusion for their state estimation. When available, commercial products tend to fuse multiple \ac{gnss} measurements for more precise state estimation and redundancy. Yet, to the best of our knowledge, \ac{eqf} formulations are not yet widely explored despite their mentioned advantages.
%even though they have several advantages compared to other techniques.

With this work, we present an implementation of a multi-position \ac{eqf} formulation and compare its performance to state-of-the-art. We will show the advantages of \ac{eqf}-based approaches for their consistency and state convergence in real-world scenarios. It will become particularly apparent that the \ac{eqf} yields good results for poorly observable states with low signal-to-noise measurements caused by the scenario and trajectory~\cite{teodori_analysis_2021, li_impact_2015}.
%\ac{mekf} implementation. We will show that the \ac{eqf} significantly outperforms the equivalent \ac{mekf} approach in consistency and state convergence in real-world situations. This is particularly apparent for poorly observable states with low signal-to-noise scenarios and trajectories.

%------------------------------------------------------------------------------%
\section{Related Work}
\label{sec:related_work}
% Filters, approaches, useability

Many current plug-and-play frameworks rely on the \ac{ekf}-based estimation and its derivatives~\cite{meier_pixhawk_2012, weiss_real-time_2011, noauthor_ardupilot_nodate, moore_generalized_2015, noauthor_px4_nodate}.

\textit{Sasiadek et al.}~\cite{sasiadek_sensor_2004} introduced the formulation of combining \ac{gnss} measurements within \iac{ins} for \acp{uav} in a \ac{kf}. While this approach presents the \ac{kf} based on the \ac{gnss} signal, they do not consider the rigid body sensor (self-)calibration or multi-\ac{gnss} sensor setups. Modern \ac{ekf}-based approaches, which went form fusing single sensor types with calibrations~\cite{weiss_versatile_2012} to extended states multi-sensor fusion~\cite{lynen_robust_2013,brommer_mars_2021}, take these into account.

% \Ac{ekf}-based approaches went from fusing single types of measurements~\cite{ssf}, to multi-measurement multi-sensor fusion~\cite{lynen_robust_2013,brommer_mars_2021}.

% While in \ac{uav} estimation research, filters have originally only fused single sensor measurements~\cite{ssf}, multi-sensor fusion was a requirement for improving state estimation~\cite{li_high-precision_2013,msf,}. Yet, all of these \ac{mekf}-based frameworks have the problem of linearizing the orientation incorrectly and thus introducing an artificial source of error into the system~\cite{whoshowed this}. Thus, in recent years also derivatives of the \ac{ekf}, the \ac{iekf}~\cite{burrau_invariant_2018} emerged.

% paragraph of eqf, why it is better than iekf, advantages, history...
Despite its success, the \ac{ekf} is known to yield poor performance and inconsistencies for high linearization error in all situations where the motion leads to unobservability of the state vector~\cite{barrau_non-linear_2015}.
To overcome issues related to high attitude linearization error and wrong initial attitude, in the eighties,
%\textit{Markley and Landis}
\textit{Lefferts et al.}~\cite{lefferts_kalman_1982, markley_attitude_2003}
explored the quaternion group as parametrization of the attitude, introducing the so-called
%\ac{mekf}~\cite{markley2003attitude},
\ac{mekf},
an algorithm that exploits the geometry of the rotational kinematics and is proven to outperform its \ac{ekf} counterpart.

Following this trend, different authors exploited the underlying symmetry of the system to design robust observers and filters. \textit{Barrau and Bonnabel}~\cite{barrau_invariant_2018, barrau_invariant_2017} introduced the \ac{iekf}, an algorithm that exploits the natural symmetry of group affine systems, and directly models the state space onto a Lie group. In the context of inertial navigation systems, the \acp{iekf} on $\SE[2]{3}$ proved superior to any \ac{ekf}-like filter in terms of performance and consistency, in particular, for \ac{gnss}-based navigation the \ac{iekf}~\cite{pavlasek_invariant_2021, barrau_navigating_2016, liu_ingvio_2023} outperform state-of-the-art solutions and to solve the inconsistencies typical to \acp{ekf}. That is, the spurious information gain, e.g., when the platform is not moving~\cite{barrau_non-linear_2015}.

In recent years the \ac{eqf} was proposed as a \emph{general filter design} for systems with symmetry~\cite{van_goor_equivariant_2020, van_goor_equivariant_2023} with specialization to the \ac{iekf} for group affine systems~\cite[Appendix B]{van_goor_equivariant_2023}.
That is, the \ac{eqf} approach is more general and applicable to a wider range of systems than the \ac{iekf} approach, as the latter one cannot properly include bias terms into the symmetry.
Thus, leveraging the extended capabilities of \acp{eqf}, \textit{Fornasier et al.}~\cite{fornasier_equivariant_2022, fornasier_overcoming_2022, fornasier_equivariant_2023} introduced a symmetry for inertial navigation systems, that properly models \ac{imu} biases, and an associated \ac{eqf} design. This work shows performance improvements compared to state-of-the-art \acp{iekf} in terms of transient behavior - meaning the convergence phase of the filter - and robustness to wrong initial conditions.

This paper builds upon the latest results in \acl{eqf} design for \acl{ins} and introduces an equivariant filter formulation for \emph{multiple} \ac{gnss} sensors, including their \emph{individual extrinsic calibration states}. Therefore, leveraging the robustness and consistency properties of equivariant filters, we design and present \iac{eqf} that can be initialized arbitrarily, requiring no knowledge about the sensor's calibration states.
%independent of the initial state estimate, thus requiring no knowledge about the sensor's calibration states.
%
% \begin{itemize}
%     \item independent of the initial state estimate, thus requiring no knowledge about the sensor's calibration states,
%     \item less error-prone given an arbitrary initial state estimate, and
%     \item more consistent than state-of-the-art commonly available \ac{mekf}-based estimators.
% \end{itemize}
% %
We will show the performance advantages of using \iac{eqf} in real-world environments by evaluating our filter of the recently published \emph{INSANE Dataset}~\cite{brommer_insane_2022}, which includes a large quantity of outdoor flights with multiple \ac{gnss} sensors and high-rate position and orientation ground-truth. Due to the lack of open-source implementations of the \ac{iekf}, we will provide a comparison to a robust, openly available implementation of \iac{mekf}~\cite{brommer_mars_2021}.

In the following sections, we first present the mathematical operators for Lie groups and its Lie algebra (\sect{sec:maths}), then introduce the biased-\ac{ins} and its symmetry (\sectt{\ref{sec:system}-\ref{sec:symmetry}}), formulate the \ac{eqf} (\sect{sec:filter_design}), and evaluate it on the \emph{INSANE Dataset}~\cite{brommer_insane_2022} (\sect{sec:evaluation}). For the interested reader, in the appendix (\appt{app:iekfdiff}), we also recall the difference between the proposed \ac{eqf} and the \ac{iekf}.

%------------------------------------------------------------------------------%
\section{Mathematical Preliminaries}
\label{sec:maths}
The following section introduces the notation and mathematical preliminaries used in this paper.
In general, bold lowercase letters are used to indicate vectors, and bold capital letters are used to indicate matrices. Non-bold letters indicate elements of a symmetry group.

$\rAB$ \diff[donates]{denotes} the translation between reference frame $\refframe[A]$ and reference frame $\refframe[B]$, $\vAB$ denotes the corresponding velocity. All vectors referenced in $\refframe[A]$ are generally \diff[donated]{denoted} by $\rA$. Further, the rotation matrix rotating a vector from $\refframe[B]$ to $\refframe[A]$ is \diff[donated]{denoted} by $\RAB$, i.e. $\rA = \RAB \rB$.

In what follows we quickly introduce some mathematical concepts about Lie groups that we used in this work. For an introduction of Lie groups theory for state estimation and robotics the reader is referred to \textit{Sol\`a et al.}~\cite{sola_micro_2021}.

Let $\vG$ be a Lie group, and $\frg$ its Lie Algebra which is isomorphic with a vector space $\R[n]$. We further define the following operators:

    % %..........................................%
    % \subsection{Lie Group}
    % \label{sec:math_lie_group}

    %..........................................%
    \paragraph{Wedge and Vee Map}
    \label{sec:math_wedge_vee}
    The wedge map is defined as a map from the vector space to the Lie algebra,
    \begin{align*}
        \group{\cdot}\W& \AtoB{\R[n]}{\frg},
        \intertext{and its inverse, the vee map, is defined as}
        \group{\cdot}\V& \AtoB{\frg}{\R[n]}.
    \end{align*}

    %..........................................%
    \paragraph{Adjoint Representations}
    \label{sec:math_adjoint}
    Given $X\in\vG$ and $\vu,\vv\in\frg$, we define the Adjoint map $\groupAdjointmap{} \AtoB{\vG \times \frg}{\frg}$ as
    \begin{align*}
        \groupAdjointmap[\gstate]{\vu\W} &= \dd\textrm{L}_{\gstate} \dd\textrm{R}_{\gstateI}[\vu\W] .
    \end{align*}
    Further, in context of Lie matrix groups we define the Adjoint matrix map as
    \begin{align*}
        \groupAdjointmatrix[\gstate]{\vu} &= \left(\groupAdjointmap[\gstate]{\vu\W}\right)\V .
    \end{align*}
    Similarly the adjoint map $\groupadjointmap{} \AtoB{\frg}{\frg}$ and the adjoint matrix map for Lie matrix groups can be defined
    \begin{align*}
        \groupadjointmap[\vu\W]{\vv\W} &= \vu\W\vv\W - \vv\W\vu\W, \\
        \groupadjointmatrix[\vu]{\vv} &= \left( \vu\W\vv\W - \vv\W\vu\W \right)\V .
    \end{align*}
    %

    % %..........................................%
    % \subsection{Group Action and Homogeneous Space}
    % \label{sec:math_group_action}

    % %..........................................%
    % \subsection{Semi-direct Product Group}
    % \label{sec:math_group_sd}

%------------------------------------------------------------------------------%
\section{Biased Inertial Navigation System}
\label{sec:system}

For this work, we consider \iac{uav} equipped with \iac{imu} measuring the robot's angular velocity and linear acceleration. Further, we consider the \ac{uav} to have $N$ uncalibrated \ac{gnss} sensors onboard, each providing global position measurements. We define $\refframe[G]$ as the global inertial frame of reference, and $\refframe[I]$ as the body-fixed \ac{imu} frame of reference. The well-known, deterministic, dynamics of the biased inertial navigation system can then be written as
%
% \begin{align}
%     \RGIdt &= \RGI ( \wI - \bwI )\W \label{equ:RGIdt},\\
%     \vGIdt &= \RGI (\aI - \baI) + \gG, \\
%     \rGIdt &= \vGIdt, \\
%     \bwIdt &= \zye, \\
%     \baIdt &= \zye. \label{equ:baIdt}
% \end{align}
\begin{IEEEeqnarray}{ll}
    \RGIdt &= \RGI ( \wI - \bwI )\W \IEEEyesnumber\IEEEyessubnumber* \label{equ:RGIdt},\\
    \vGIdt &= \RGI (\aI - \baI) + \gG, \\
    \rGIdt &= \vGIdt, \\
    \bwIdt &= \zye, \\
    \baIdt &= \zye. \label{equ:baIdt}
\end{IEEEeqnarray}
$\RGI$ refers to the rotation of the \ac{imu}, and $\rGI$ and $\vGI$ correspond to the \ac{imu} position and velocity in the global frame $\refframe[G]$. $\wI$ and $\aI$ are the body-fixed, biased inputs from the \ac{imu} and refer to the angular velocity and linear acceleration, respectively, with its biases defined as $\bwI$ and $\baI$. $\gG = \bmat{0,&0,&9.81}\T\si{\meter\per\second\squared}$ refers to the gravity vector expressed in the global frame.

Let the core state space be defined as $\cC = \SO \times \R[12]$ % \times \R[3] \times \R[3] \times \R[3]$
and the core state be defined as
\begin{align*}
    \core &= \left( \RGI, \vGI, \rGI, \bwI, \baI \right) \in \cC .
\end{align*}
Let a zero-dynamic individual calibration state be defined as
\begin{align*}
    \calib[i] &= \left( \rIP[i] \right) \inR[3],
\end{align*}
with
\begin{align}
    \rIPdt[i] &= \zye .\label{equ:rIPdt}
\end{align}
Then the full state in the state space $\cM = \cC \times (\R[3])^{N}$ can be defined as
\begin{align*}
    \state &= \left( \core, \calib[1], \dots, \calib[N] \right) \in \cM .
\end{align*}

We define the system's state space as $\state = \group{\TGI, \bI, \tIP[1], \dots, \tIP[N]} \in \cM$, where $\TGI = \group{\RGI, \vGI, \rGI} \in \Se[2]{3}$ \diff[donates]{denotes} the \ac{uav}'s extended pose, $\bI = \group{\bwI, \baI} \inR[6]$ the \ac{imu}'s biases, and $\tIP[i] = \rIP[i] \inR[3]$ the $i$-th sensor's calibration state. Further, $\sysin = \group{\wI, \aI} \in \bbL \subseteq\R[6]$ describes the system's input.
To simplify the notation in the subsequent sections the frame indices $\refframe[G]$, $\refframe[I]$, and $\refframe[P_i]$ are omitted.

% renew commands
\regenerateRoboticSymbols{rGI}{pos}[][][]
\regenerateRoboticSymbols{vGI}{linvel}[][][]
\regenerateRoboticSymbols{aGI}{linacc}[][][]
\regenerateRoboticSymbols{qGI}{quat}[][][]
\regenerateRoboticSymbols{RGI}{romat}[][][]
\regenerateRoboticSymbols{gG}{gravity}[]
\regenerateRoboticSymbols{tIP}[O{}]{trans}[][][#2]
\regenerateRoboticSymbols{aI}{linacc}[][][]
\regenerateRoboticSymbols{wI}{angvel}[][][]
\regenerateRoboticSymbols{baI}{bias}[][\va]
\regenerateRoboticSymbols{bwI}{bias}[][\omega]

Then the system described in \equst{\ref{equ:RGIdt}-\ref{equ:baIdt}} and \equt{equ:rIPdt} can be rewritten as
%
% \begin{align}
%     \TGIdt &= \TGI \left( \WI - \BI + \Dir \right) + \left( \GG - \Dir \right)\TGI \label{equ:TGIdt} \\
%     \bIdt &= \zye \label{equ:bIdt}\\
%     \tIPdt[i] &= \zye ,\label{equ:tIPidt}
% \end{align}
\begin{IEEEeqnarray}{l l}
    \TGIdt &= \TGI \left( \WI - \BI + \Dir \right) + \left( \GG - \Dir \right)\TGI \IEEEyesnumber  \IEEEyessubnumber* \label{equ:TGIdt} \\
    \bIdt &= \zye \label{equ:bIdt}\\
    \tIPdt[i] &= \zye ,\label{equ:tIPidt}
\end{IEEEeqnarray}
where
\begin{align*}
    \WI &= \bmat{\wI{}\W & \aI & \zye[3][1]\\ \multicolumn{3}{c}{\zye[2][5]}} \in \se[2]{3} \subset \R[5][5],\\
    \BI &= \bmat{\bwI{}\W & \baI & \zye[3][1]\\ \multicolumn{3}{c}{\zye[2][5]}} \in \se[2]{3} \subset \R[5][5],\\
    \GG &= \bmat{\zye[3][3] & \gG & \zye[3][1]\\ \multicolumn{3}{c}{\zye[2][5]}} \in \se[2]{3} \subset \R[5][5], \text{ and}\\
    \Dir &= \bmat{\zye[3][4]& 0\\ \zye[1][4] & 1\\ \zye[1][4] & 0} \in \R[5][5] .
\end{align*}

To simplify notation, in the following sections, the elements related to the position sensors (calibrations) are denoted only once with index $i$. Please note that for $N$ sensors, $1\leq i\leq N$, these elements would be repeated $N$ times.
%------------------------------------------------------------------------------%
\section{Semi-direct Product Symmetry}
\label{sec:symmetry}

% This section proposes and discusses an extended symmetry of our previous work~\cite{TRO}. This symmetry is based on the geometrical structure given by a semi-direct product group $\sdgrp$ and does not introduce any additional virtual states compared to other works~\cite{fornasier_equivariant_2022}.

In this section, we present the symmetry of the previously defined system. The symmetry is based on the semi-direct product group $\sdgrp$~\cite{ng_attitude_2019, ng_pose_2020} and builds upon the author's previous work~\cite{fornasier_equivariant_2022, fornasier_equivariant_2023}.
%and does not introduce any additional virtual states compared to other works~\cite{fornasier_equivariant_2022}.

Given the revised system dynamics in \equtt{\ref{equ:TGIdt}-\ref{equ:tIPidt}}, we define the symmetry group to be $\vG := (\sdgrp) \ltimes (\R[3])^N$. Let $\gstate = \group{\grpC, \grpc , \grpd[i]} \in \vG$ be an element of the symmetry group, with $\grpC = \group{\grpB, \grpb} \in \SE[2]{3}$ and $\grpB = \group{\grpA, \grpa} \in \SE{3}$. Then the inverse element is given by $\gstateI = \group{\grpCI, \smin \groupAdjointmap[\grpBI]{\grpc}, \smin \grpAT\grpd[i]}$ with $\grpCI = \group{\grpAT, \smin\grpAT{\grpa}, \smin\grpAT{\grpb}}$, and the identity element of $\vG$ is $\id = \group{\eye[3], \zye, \zye, \zye, \zye} \in \vG$.

    %..........................................%
    \subsection{Equivariance}
    \label{sec:equi}
    This section presents \emph{the group action} and \emph{equivariant configuration output} required for the later \ac{eqf} design.

    % \begin{lemma}
    %     Define $\gaction \AtoB{\vG \times \cM}{\cM}$ as
    %     %
    %     \begin{align}
    %         \groupaction{\gstate,\state} &:= \group{\vT\grpC, \groupAdjontmatrix[\grpBI]{\bI - \grpcV}, \grpAT(\rIP[i] - \grpd[i])} \in \cM
    %     \end{align}
    %     %
    %     Then $\gaction$ is a transitive right group action of G on $\cM$.
    % \end{lemma}
    \begin{definition}
        The transitive right group action of $\vG$ on $\cM$, $\gaction \AtoB{\vG \times \cM}{\cM}$ for our filter is defined as
        \begin{align}
            \groupaction{\gstate,\state} &:= \group{\TGI\grpC, \groupAdjointmatrix[\grpBI]{\bI - \grpcV}, \grpAT(\tIP[i] - \grpd[i])} \in \cM .
        \end{align}
    \end{definition}

    % proof?

    % \begin{lemma}
    %     We define $h_i(\state) \in \cN$ a measurement of a known vector quantity $\vde$ as follows
    %     %
    %     \begin{align}
    %         h_i(\state) &= \RGIT (\vde_i - (\rGI + \RGI\rIP[i])) \label{equ:pos_meas}
    %     \end{align}
    %     %
    %     for $i=1,\dots,N$. Let $\sysout[i] = h_i(\state) \in N$ be a measurement defined according to the above model. Then, the configuration output $\cfgout \AtoB{G\times \cN}{\cN}$ defined as
    %     %
    %     \begin{align}
    %         \confoutput[i]{\gstate,\sysout[i]} &:= \grpAT (\sysout[i] - \grpb + \grpd[i])
    %     \end{align}
    %     %
    %     is equivariant.
    % \end{lemma}
    % %
    % We take leverage of our previous work~\cite{TRO} to remodel uncalibrated position sensor measurements according to \equt{equ:pos_meas} and construct the $\vde_i$ vector with the $i$-th sensor's raw measurements and assuming $\sysout[i] = \zye$.

    \begin{definition}
        Given the defined measurement equation $\sysout[i] = h_i(\state) \in \cN$ of a known vector quantity $\vde_i$ with~\cite{fornasier_equivariant_2023}
        \begin{align}
            h_i(\state) &= \RGIT (\vde_i - (\rGI + \RGI\tIP[i])), \label{equ:pos_meas} %\RGI\rIP[i]
        \end{align}
        for $1\leq i \leq N$.
        Then the equivariant configuration output of our filter, $\cfgout \AtoB{\vG\times \cN}{\cN}$ can be defined as
        \begin{align}
            \confoutput[i]{\gstate,\sysout[i]} &:= \grpAT (\sysout[i] - \grpb + \grpd[i]) .
        \end{align}
    \end{definition}

    In the \ac{eqf} design below, the vector $\vde_i$ is set to zero.%, i.e. $\vde_i = \zye$.

    %..........................................%
    \subsection{Equivariant Lift}
    \label{sec:lift}

    %To design an equivariant filter, the system input needs to be lifted~\cite{mahony_equivariant_2020-1}.
    In the context of equivariant filtering \emph{the lift}~\cite{mahony_equivariant_2020-1} provides the structure that connects the input of the system posed on the homogeneous space to the input of the system lifted onto the symmetry group. For the given system in~\equtt{\ref{equ:TGIdt}-\ref{equ:tIPidt}} the lift is defined as follows.
    %We leverage the fact that the existence of a transitive group action of the symmetry group $G$ on the state space $\cM$ guarantees the existence of a Lift \cite{mahony_equivariant_2020-1,TRO}.
    %
    % \begin{theorem}
    %     Define $\lft[1] \AtoB{\cM\times L}{\se[2]{3}}$, $\lft[2] \AtoB{\cM\times L}{\se{3}}$, and $\lft[i] \AtoB{\cM\times L}{\R[3]}$ as
    %     %
    %     \begin{align}
    %         \lift[I]{\state,\sysin} &:= (\WI - \BI + \Dir) + \TGIT(\GG - \Dir)\TGI, \\
    %         \lift[II]{\state,\sysin} &:= \\
    %         \lift[i]{\state,\sysin} &:= - (\wI - \bwI)\W \ \rIP[i] \quad i=1,...,N.
    %     \end{align}
    %     %
    %     with $map from TRO$. Then the map $\lift{\state,\sysin} = \group{\lift[I]{\state,\sysin}, \lift[II]{\state,\sysin}, \lift[i]{\state,\sysin}}$ is a lift for system defined in XX.
    % \end{theorem}
    %
    \begin{definition}
        We define the map ${\lift{\state,\sysin} = \group{\lift[I]{\state,\sysin}, \lift[II]{\state,\sysin}, \lift[i]{\state,\sysin}}}$ as a lift for the system defined in \equst{\ref{equ:TGIdt}-\ref{equ:tIPidt}} with
        %
        % \begin{align}
        %     \lift[I]{\state,\sysin} &:= (\WI - \BI + \Dir) + \TGII(\GG - \Dir)\TGI \\
        %     \lift[II]{\state,\sysin} &:= \groupadjointmap[\bIW]{\Pi(\lift[I]{\state,\sysin})} \\
        %     \lift[i]{\state,\sysin} &:= - (\wI - \bwI)\W \ \tIP[i] \quad i=1,...,N
        % \end{align}
        \begin{IEEEeqnarray}{ll}
            \lift[I]{\state,\sysin} &:= (\WI - \BI + \Dir) + \TGII(\GG - \Dir)\TGI \IEEEyesnumber\IEEEyessubnumber* \\
            \lift[II]{\state,\sysin} &:= \groupadjointmap[\bIW]{\Pi(\lift[I]{\state,\sysin})} \\
            \lift[i]{\state,\sysin} &:= - (\wI - \bwI)\W \ \tIP[i] \quad i=1,...,N
        \end{IEEEeqnarray}
        where $\lft[1] \AtoB{\cM\times \bbL}{\se[2]{3}}$, $\lft[2] \AtoB{\cM\times \bbL}{\se{3}}$, and $\lft[i] \AtoB{\cM\times \bbL}{\R[3]}$.
        $\Pi \AtoB{\se[2]{3}}{\se{3}}$ is a map defined such that, \diff{${\forall \vx,\vy,\vz \inR[3] \vert (\vx,\vy,\vz) \inR[9]}$, ${\Pi((\vx,\vy,\vz)\W) = (\vx,\vy)\W \in \se{3}}$}.
    \end{definition}

%------------------------------------------------------------------------------%
\section{Equivariant Filter Design}
\label{sec:filter_design}
We extend the filter design of our previous work \cite{fornasier_equivariant_2022,fornasier_overcoming_2022, fornasier_equivariant_2023} with the presented symmetries given measurements from \iac{imu} propagation sensor and $N$ \ac{gnss} position update sensors. Similar to our previous work, we initialize the filter to the origin of the state space $\state<0> = \id$.

We further define local coordinates $\err$ for the equivariant error ${e := \groupaction{\gstateI<est>, \state}}$, to be normal coordinates, thus ${\err = \errorfx{e} := \log(\groupactionI[\state<0>]{e})\V \inR[n]}$, where ${\log\AtoB{\vG}{\frg}}$ is the logarithm of the symmetry group. Note that given the \ac{eqf} estimate in the symmetry group at time $t$, that is ${\gstate<est>(t)}$, the system's estimate at time $t$ is written ${\state<est>(t) = \groupaction{\gstate<est>(t),\state<0>}}$~\cite{van_goor_equivariant_2023}.

We can derive the linearized error state, and output matrices
\begin{align}
    \errdt &\simeq \statematZ[t] \err, \\
    \delta\left( h(e) \right) &= \delta\left( \confoutput{\gstateI<est>, h(\state)} \right) \approx \outmatS[t] \err, \label{equ:outlin}
\end{align}
where $\statematZ[t]$ and $\outmatS[t]$ are defined according to~\cite{fornasier_equivariant_2023}.
%
% \begin{align}
%     \statematZ[t] &= D .... \\ % eq 51
%     \outmatC[t] \err &= \frac{1}{2} \left( D...  + D... \right) Ad %eq 35
% \end{align}

When formulating filters for real-world data, one cannot assume all sensors' measurements to be available simultaneously. The following presents a general filter formulation applying updates for the $i$-th position sensor. This formulation allows an arbitrary number of sensor updates at different rates, as in real-world environments.

Let $\err = \errorfx{e} = \group{\err[\RGI], \err[\vGI], \err[\rGI], \err[\bI], \err[\tIP[i]]} \inR[18]$ and using the previous equations, we can then derive the linearized filter matrices for the $i$-th sensor update as
\begin{align}
    \statematZ[t] &= \bmat{
        \vUp_1 & \vUp_2 & \zye[9][3N] \\
        \zye[6][9] & \vUp_3 & \zye[6][3N] \\
        \zye[3N][9] & \zye[3N][6] & \vUp_4
    } \inR[(15+3N)][(15+3N)] \label{equ:At}\\
    \outmatS[t,i] &= \bmat{
        \vUp_5 & \zye[3][3] & \smin\eye[3] & \zye[3][6] & \vUp_6
    } \inR[3][(15+3N)]
\end{align}
where,
\begin{align*}
    \vUp_1 &= \bmat{
        \zye[3][3] & \zye[3][3] & \zye[3][3] \\
        \gGW & \zye[3][3] & \zye[3][3] \\
        \zye[3][3] & \eye[3] & \zye[3][3]
    } \inR[9][9], \\
    \vUp_2 &= \bmat{
        \multicolumn{2}{c}{\eye[6]}\\
        \grpbW<est> & \zye[3][3]
    } \inR[9][6], \\
    \vUp_3 &= \groupadjointmatrix[\vUp_7]{} \inR[6][6], \\%\groupadjointmatrix[[\groupAdjointmatrix[\BI<est>]{\wI + \grpcV<est>} + \GG]]{} \inR[6][6], \\
    \vUp_4 &= \diag(\vGa_1, \dots, \vGa_N) \inR[3N][3N], \\
    \vUp_5 &= \frac{1}{2}(\sysout[i]  + \grpb<est> - \grpd<est>[i])\W \inR[3][3],\\
    \vUp_6 &= \bmat{\delta_{1,i}\eye[3], & \dots, & \delta_{N,i}\eye[3]} \inR[3][3N], \\
    \vUp_7 &= \groupAdjointmatrix[\grpB<est>]{\wI} + \grpcV<est> + \GG<avg> \in \se[2]{3} \subset \R[4][4], \text{ and} \\
    \vGa_1 &= \dots = \vGa_n = (\grpA<est>\,\wI + \grpcV<est>[\omega])\W \inR[3][3].
\end{align*}
$\delta_{j,i}$ refers to the Kronecker delta, which is $1$ if $i=j$ and $0$ otherwise. Thus, $\vUp_6$ is a $3$-by-$3N$ zero matrix, except the $i$-th $3$-by-$3$ block being the identity matrix, corresponding to the $i$-th sensor update. $\GG<avg> \in\se{3} \subset \R[4][4]$ refers to the upper-left 4-by-4 matrix of $\GG$.

Finally, our discrete-time \ac{eqf}'s propagation and update equations are quite similar to the one of \iac{ekf}~\cite{fornasier_overcoming_2022}.
%
% \begin{align}
%     % &\covmatprior[k+1] = \Phi \covmatpost[k] \Phi\T + MD \dt
%     &\innocov = \outmatC[t,i]\covmatprior[k+1]\outmatC[t,i]{}\T + \vN, \IEEEyesnumber\IEEEyessubnumber*\\
%     &\ekfgain = \covmatprior[k+1] \outmatC[t,i]{}\T \innocovI, \\
%     &\delta = \confoutput[i]{\gstateI<est>,0} - \sysout[i] \\
%     &\innores = \frechetderivative{E}{\id}[\groupaction{\state,E}] \dd\errI \cdot \ekfgain \cdot \delta \\ %<0>
%     % \confoutput[i]{\gstateI<est>,\sysout[i]}, \\
%     &\gstatepost<est>[k+1] = \exp(\innores) \gstateprior<est>[k+1],  \\
%     &\covmatpost[k+1] = (\eye - \ekfgain\outmatC[t,i])\covmatprior[k+1],
% \end{align}
\begin{IEEEeqnarray}{l}
    \innocov = \outmatS[t,i]\covmatprior[k+1]\outmatS[t,i]{}\T + \vN, \IEEEyesnumber\IEEEyessubnumber*\\
    \ekfgain = \covmatprior[k+1] \outmatS[t,i]{}\T \innocovI, \\
    \delta = \confoutput[i]{\gstateI<est>,0} - \sysout[i] \\
    \innores = \frechetderivative{E}{\id}[\groupaction{\state<0>,E}] \dd\errfvalI \cdot \ekfgain \delta \\
    \gstatepost<est>[k+1] = \exp(\innores) \gstateprior<est>[k+1],  \\
    \covmatpost[k+1] = (\eye - \ekfgain\outmatS[t,i])\covmatprior[k+1],
\end{IEEEeqnarray}
 where $\vN \in \bbS_{+}(3) \subset \R[3][3]$ is an output gain matrix, and $\covmat \in \bbS_{+}(15+3N) \subset \R[15+3N][15+3N]$ the Riccati or covariance matrix (of the error in local coordinates). $\gstate<est> \in \vG$ is the \ac{eqf} state with the initial state $\gstate<est>(0) = \id$. The main difference to an \ac{ekf} formulation is the definition of the innovation residual $\innores$, where ${\frechetderivative{E}{\id}[\groupaction{\state<0>,E}] \dd\errfvalI}$ maps the scaled residual
 % ${\ekfgain \cdot \confoutput[i]{\gstateI<est>,\sysout[i]}}$
 ${\ekfgain \delta}$
 into the Lie algebra of the symmetry group.

%------------------------------------------------------------------------------%
\section{Experimental Evaluation}
\label{sec:experiments}

To evaluate our filter design and show the usability of real data, we use the \emph{INSANE Dataset}~\cite{brommer_insane_2022}. This dataset provides recordings with over 14 sensors, including two \ac{gnss} sensors offset by approx. \SI{1}{\meter} from the \ac{imu}. To show the usability of the proposed filter, we use the 19 ``\textit{Mars}'' (outdoor) and the ``\textit{Outdoor}'' recordings of the \emph{INSANE Dataset}.
%as he did in.
We select this dataset as it includes a wide range of \ac{uav} flights, from long-distance to short pickup-and-place, high-speed to low-speed flights. It also contains an outdoor orientation ground-truth, which is derived geometrically from raw sensor measurements using the vehicle's geometry and sensors' calibration.
%Thus, the ground-truth contains the sensors' noise and we expect any filter framework to smoothen this trajectory - but also rarely yields no error due to the noise.

Note that, due to external influences in this dataset not all flights have a highly accurate \acs{rtk}-\ac{gnss} fix available. Since these measurements are also used to generate the ground-truth orientation, slight position measurement errors can yield larger orientation errors, rendering the ``ground-truth'' less accurate. The affected datasets are marked with a star ($*$) in \tabt{tab:rmse}.

    %..........................................%
    \subsection{Comparison to \acs{mekf} approaches}
    \label{sec:ekf_comparison}

    Due to the lack of open-source implementation of \iac{iekf} - with neither single nor multiple-\ac{gnss} sensors - we choose to compare our presented filter to an implementation of a state-of-the-art modular \ac{mekf}~\cite{brommer_mars_2021} to have at least well-known comparison data. This \ac{mekf} implementation directly provides (multiple) \ac{gnss} sensor modules with very little modification. For a detailed performance analysis of the consistency and core error state performance we refer the reader to our previous work~\cite{fornasier_equivariant_2023}.
    We further want to highlight that the \ac{mekf} formulation is able to handle asynchronous and out-of-order measurements, while the \ac{eqf} implementation is designed in a simpler way to show the proof of concept.

    % To show the advantages of \ac{eqf}-based approaches we compare our framework to an implementation of a state-of-the-art \ac{mekf}.
    % probably needed, as a proper eqf implementation would improve even more
    % We want to highlight that our \ac{mekf} formulation is able to handle asynchronous and out-of-order measurements, while the \ac{eqf} implementation is designed in a simpler way to show the proof of concept.

    We initialize both filters with the same initial state, i.e., $\state<0> = \id = \group{\eye[3], \zye, \zye, \zye, \zye, \zye, \zye} \in \cM$. We want to highlight the purposely wrong chosen initial state for both \ac{gnss} sensor calibrations, whose ground-truth values are
    \begin{align*}
        \tIP[1] &= \bmat{0.35, 0.41, 0}\T \si{\meter} \\
        \tIP[2] &= \bmat{-0.47, -0.41, 0}\T \si{\meter}.
    \end{align*}
    %
    % ($\approx\pm\SI{20}{\percent}$) <-- cannot find ref anymore, so won't put it for now
    Given that this initial estimate is most likely outside the beacon of convergence for the \ac{mekf}, we acknowledge that \ac{ekf}-based approaches tend to require an initial state within the proximity off the ground-truth. However, with the wrong initial state estimate, we want to highlight the performance of symmetry-based approaches concerning the convergence and rate.
    % improvement of \ac{eqf}-based approaches concerning the convergence (rate).

    Further, both filters are initialized with the same state covariance $\covmat[0]$ and use the same measurement noise.

    \begin{comment}
    %..........................................%
    \subsection{Simulation}
    \label{sec:simulation}

    To verify our approach we perform 100 simulations with randomly generated trajectories. The duration of each trajectory is lasting \SI{60}{\second}, with \ac{imu} and \ac{gnss} sensor rates set to \SI{200}{\hertz} and \SI{8}{\hertz}, respectively.

    % Avg RMSE:
    %     >> mean(rmses.est.p)
    % ans =
    %     0.1254
    % >> mean(rmses.est.a)
    % ans =
    %     0.6276
    % >> mean(rmses.mars.p)
    % ans =
    %     5.0115
    % >> mean(rmses.mars.a)
    % ans =
    %    77.0822

    \end{comment}

    %..........................................%
    \subsection{Evaluation Results}
    \label{sec:evaluation}

\sisetup{detect-weight=true,detect-inline-weight=math}
\begin{table}[t]
    \footnotesize
    \caption{\Acs{rmse} evaluation in the asymptotic phase of our \acs{eqf} framework compared to \iacs{mekf}-based implementation on the \emph{INSANE Dataset}~\cite{brommer_insane_2022}.}%
    \label{tab:rmse}
    \begin{center}
        \begin{tabular}{ l l @{} S[table-format = 1.4] S[table-format = 2.4] S[table-format = 1.4] S[table-format = 3.4] @{} }
        \toprule %\hline
        \multicolumn{2}{c}{\textbf{ Dataset}} & \multicolumn{2}{c}{\textbf{ \acs{eqf} \acs{rmse}}} & \multicolumn{2}{c}{\textbf{ \acs{mekf} \acs{rmse} Pos.}}\\
        No. & $t_0 [\si{\second}]$& \textbf{ Pos. [\si{\meter}]} & \textbf{ Att. [\si{\degree}]} & \textbf{ Pos. [\si{\meter}]} & \textbf{ Att. [\si{\degree}]}\\
        \midrule
        M01 & 20 &  \bfseries 0.1089 & 12.7486& 0.5264 & \bfseries 2.5468 \\
        M02 & 23 &  \bfseries 0.0890 & \bfseries 3.8077 & 4.2120 & 128.5142 \\
        M03 & 29 &  \bfseries 0.1529 & \bfseries 2.5925 & 5.6688 & 97.5796 \\
        M04$*$ & 34 &  \bfseries 0.3063 & \bfseries 5.7552 & 2.4872 & 45.7479 \\
        M05 & 50 &  \bfseries 0.1449 & \bfseries 1.9385 & 1.1497 & 8.0534 \\ % run_old
        M06 & 25 &  \bfseries 0.0966 & \bfseries 3.3420 & 0.1237 & 7.4513 \\ % run_old
        M07 & 36 &  \bfseries 0.1766 & \bfseries 2.7015 & 0.6316 & 11.5197 \\ % run_old
        M08$*$ & 43 &  \bfseries 0.2205 & \bfseries 9.0050 & 0.3291 & 11.9717 \\ % run_old
        M09$*$ & 34 &  \bfseries 0.4682 & \bfseries 10.8112 & 3.6165 & 38.2228 \\
        M10 & 42 &  \bfseries 0.2332 & \bfseries 6.4940 & 1.2518 & 12.9468 \\ % noise4
        M11 & 31 &  \bfseries 0.3415 & \bfseries 5.7310 & 2.4224 & 36.0030 \\ % run_old
        M12 & 27 &  \bfseries 0.3705 & \bfseries 2.8902 & 1.7011 & 10.4460 \\
        M13 & 10 &  \bfseries 0.1957 & \bfseries 3.1973 & 2.7391 & 76.1869 \\
        M14$*$ & 45 &  \bfseries 0.8265 & \bfseries 28.5057 & 2.7604 & 60.5424 \\ % noise4
        M15 & 14 &  \bfseries 0.2452 & \bfseries 4.1148 & 2.3284 & 30.6290 \\
        M16$*$ & 42 &  \bfseries 0.1955 & \bfseries 6.9932 & 0.2311 & 6.8966\\
        M17$*$ & 35 &  \bfseries 0.1660 & \bfseries 6.9044 & 3.1691 & 50.2471 \\
        M18$*$ & 54 &  \bfseries 0.1289 & \bfseries 9.0023 & 0.5740 & 19.1046 \\
        M19$*$ & 23 &  \bfseries 0.1217 & \bfseries 4.3011 & 1.3056 & 6.3813 \\
        % \midrule
        O01$*$ & 50 &  \bfseries 0.2251 & \bfseries 11.3116 & 0.8893 & 17.4661 \\ % run_old
        \bottomrule
        \end{tabular}
    \end{center}
\end{table}

    We evaluate the 20 above-mentioned recordings of the \emph{INSANE Dataset} on both filter frameworks.
    Due to the long static duration at the beginning of each recording and the known observability/consistency issues for \ac{ekf}-based frameworks, an empirically derived starting time $t_0$ was chosen for each evaluation (c.f. \tabt{tab:rmse}) in favor of the \ac{mekf} approach.
    As it is known in \ac{ekf}-based approaches, noisy position measurements from off-center sensors can lead to incorrect yaw estimation when there is no motion present. Thus, for a fair comparison, we want to limit this effect only to the motion-full part of the dataset.

    The \ac{rmse} of each individual evaluation is given in \tabt{tab:rmse}. As expected, the \ac{eqf} performs well in most datasets compared to \ac{mekf} approaches.
    % As a result of the consistent filter derivation and better orientation (error) representation, the \ac{eqf} yields good results throughout the datasets.
    %
    % The \ac{eqf} outperforms the \ac{mekf} framework in most datasets, especially regarding the orientation \ac{rmse}. This is a result of the filter derivation and better orientation (error) representations compared to \ac{ekf}-based formulations.
    %
    We can further show that, without any knowledge of the sensor calibration, i.e., initializing these states at identity, the \ac{eqf} performs well for all flights of the \emph{INSANE Dataset}.
    % This shows, that the \ac{eqf}'s convergence property makes it a suitable state estimator for modular sensor modules on \acp{uav}.
    For interested readers, we also provide the trajectory comparison of all runs in \appt{app:insane_evaluation}.
    % rest of paragraph on aTE table, might change if started just before

    % General comparison
    % In \figt{fig:results_insane}, the results of the proposed \ac{eqf} for the 20 datasets of \emph{INSANE} are presented and compared to the ground truth. It is clearly visible that our framework can correctly estimate the vehicle's position in different flight scenarios.
    % I should put a table with the ATE here! <-- is currently being calculated

    % \newcommand{\plotrun}{runs_230301}
    \newcommand{\plotrun}{runs_icar}
    \newcommand{\plotcompold}{mars_5}
    \newcommand{\plotcomp}{mars_5_final}
    \newcommand{\runnum}{5}
    \newcommand{\runshort}{M0\runnum}
    \newcommand{\plotfiletype}{pdf}

\begin{figure}
        \centering
        % \subfloat[Position error \label{fig:res_detailed_pos}]{
            \includegraphics[width=\linewidth]{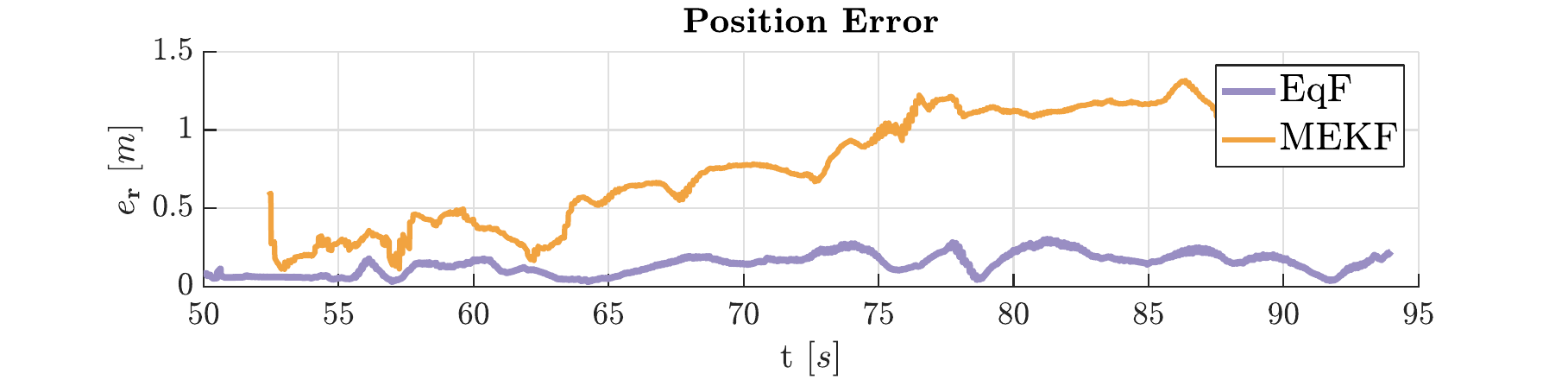} \\%
        % } \\
        % \subfloat[Orientation error \label{fig:res_detailed_ori}]{
            \includegraphics[width=\linewidth]{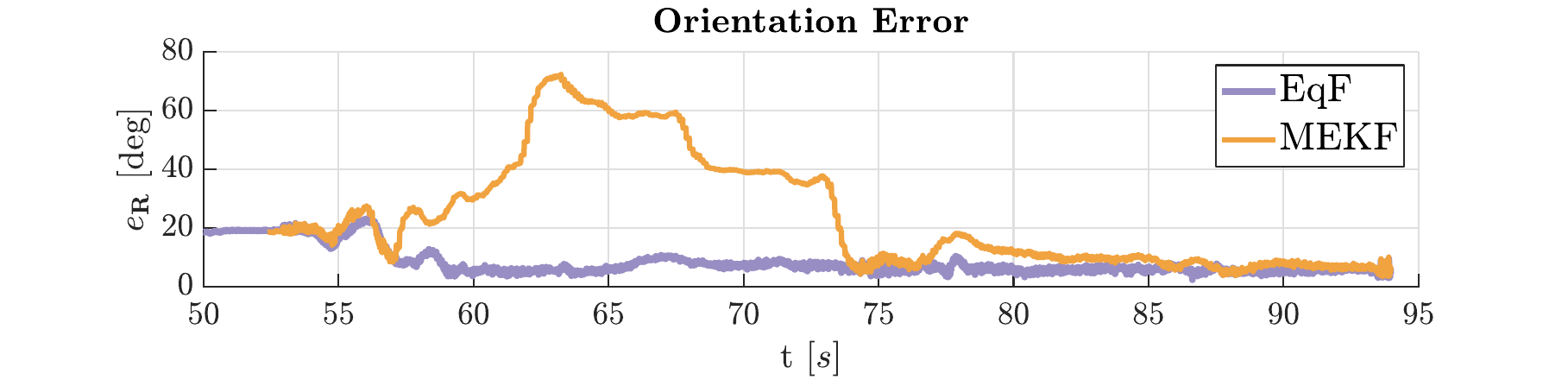}%
        % }
        \vspace{-0.3cm}
        \caption{Pose error for the \acs{eqf} and \acs{mekf} frameworks on dataset \runshort.}
        \label{fig:res_detailed_pose}
        \vspace{-0.5cm}
    \end{figure}

\begin{figure}
        \centering
        % \subfloat[Calibration States \label{fig:res_detailed_calib}]{
            \includegraphics[width=\linewidth]{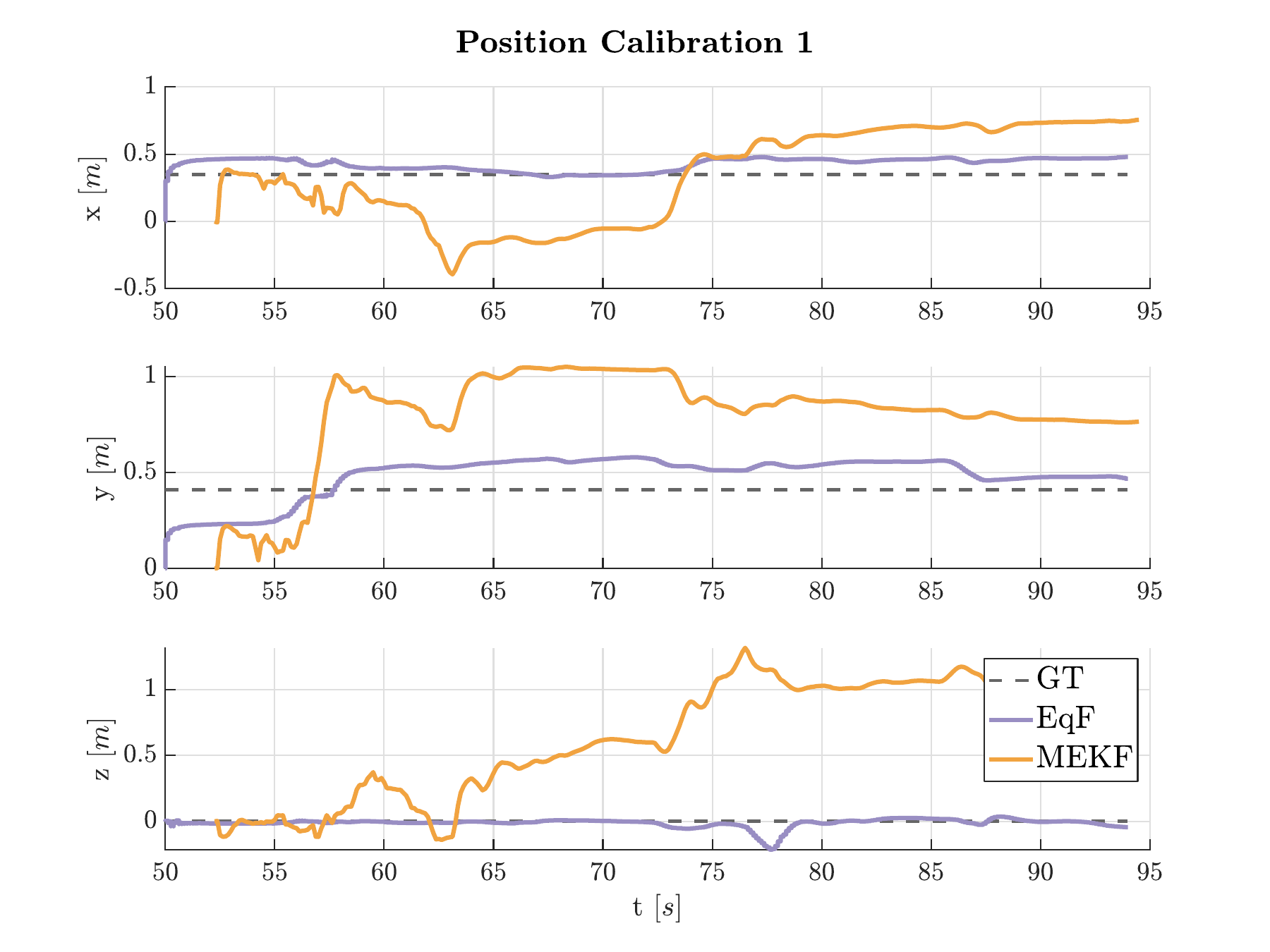} \\%
            \vspace{-0.21cm}
            \includegraphics[width=\linewidth]{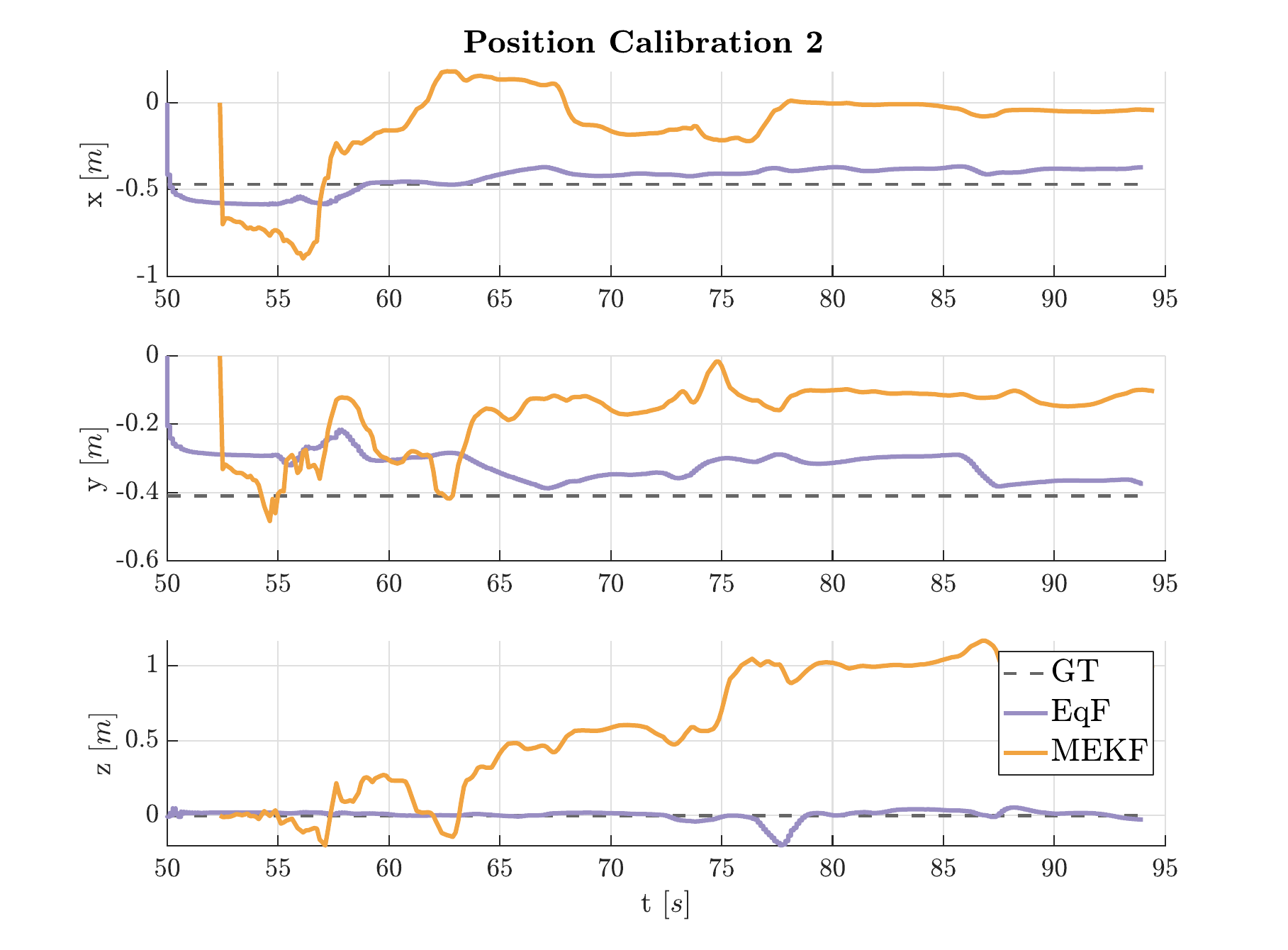}
        % } \\
        \vspace{-1cm}
        \caption{Calibration states comparison between the \acs{eqf} and \acs{mekf} frameworks and ground-truth (``GT'') on flight \runshort. Both frameworks are intentionally initialized with a wrong calibration, $\tIP<est>[1](t_0)=\tIP<est>[2](t_0)=\zye$. Yet, the \acs{eqf} framework can converge to the correct states (faster) in contrast to the \acs{mekf}-based approach.}
        \label{fig:res_detailed_calib}
        \vspace{-0.5cm}
    \end{figure}

\begin{figure}
        \centering
        % \subfloat[Position error \label{fig:res_detailed_pos}]{
            \includegraphics[width=\linewidth]{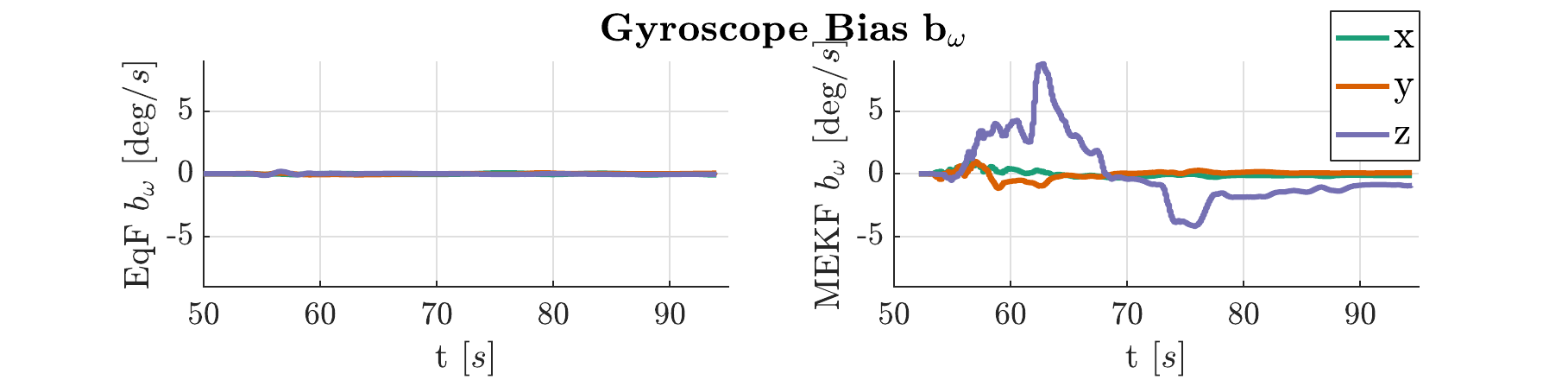} \\%
        % } \\
        % \subfloat[Orientation error \label{fig:res_detailed_ori}]{
            \includegraphics[width=\linewidth]{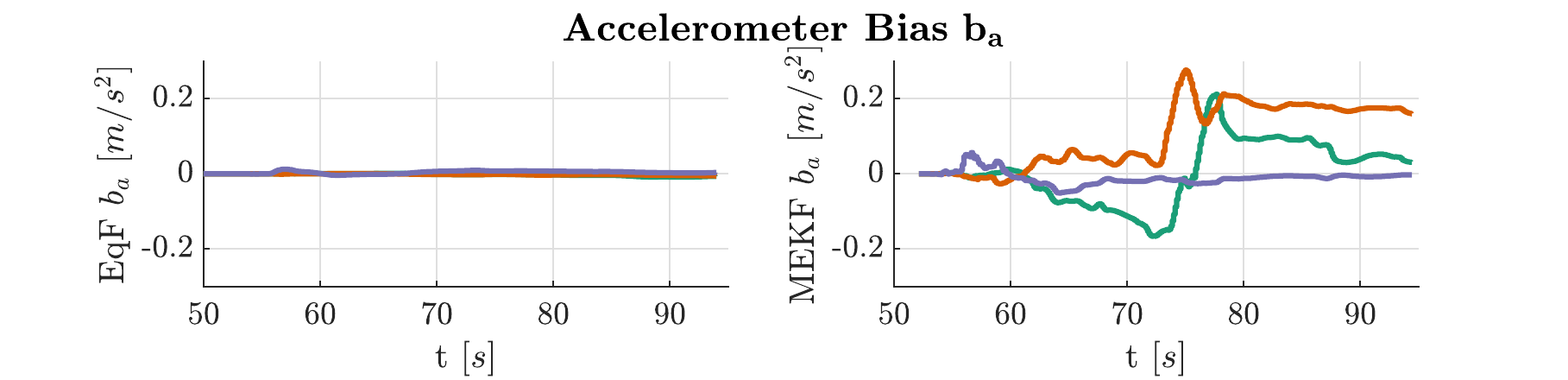}%
        % }
        \vspace{-0.3cm}
        \caption{\add{Bias estimates of the \acs{eqf} and \acs{mekf} frameworks on dataset \runshort.}}
        \label{fig:res_bias}
    \end{figure}

    % selected run, e.g. INSANE.mars18 (lots of takeoff and land?)
    Furthermore, for the selected dataset \textit{Mars \runnum ~(\runshort)}, we compare the pose errors for the \ac{eqf} and \ac{mekf} implementation in \figt{fig:res_detailed_pose} with the calibration state \add{and bias} estimates displayed in \figt{fig:res_detailed_calib} \add{and \figt{fig:res_bias}, respectively}.
    We choose this dataset as it represents a low signal-to-noise ratio scenario with generally little excitation and thus decreased yaw observability throughout the entire run.
    % This in turn lowers the impact of the noise on the measurements used to generate the ground-truth for this dataset. Also it has a highly accurate \ac{rtk}-\ac{gnss} fix.
    %has little yaw motion while traversing forward and backward, thus generally decreasing the yaw observability for the given setup and without prior calibration knowledge.
    % While in this run the \ac{eqf} accumulates a minor error in the position estimation due to an imperfect calibration convergence, it stays constant throughout the flight similar to an initial drift in \ac{vio} frameworks. Yet, the \ac{mekf}'s calibration states are not converging (correctly) and thus the position error is larger and fluctuates.


    In this comparison, we can clearly see the advantages of the \ac{eqf} framework. Both frameworks are able to converge to a good orientation estimate, though the \ac{mekf} has a noticeable longer convergence time. However, as can be expected, while improving on the orientation estimate the \ac{mekf} trades the position accuracy as its calibration states are not converging correctly.
    %hile both frameworks are able to converge at a good orientation estimate (the \ac{mekf} only after noticeably convergence time), as can be expected, the \ac{mekf} trades the position accuracy for it as the calibration states are not converging correctly. Especially the $z$-calibration states and position estimates are erroneous.
    The \ac{eqf} converges fast and correctly to the actual sensor calibrations. As a result, the \ac{eqf} is able to provide accurate state estimates even if the rigid body sensor calibrations are initially completely unknown.

    For the sake of completeness, we acknowledge and tested the \ac{mekf} with initial states only \SI{5}{\percent} off ground-truth values and \add{we} were able to achieve similar estimation results as the \ac{eqf}. Naturally, the low initial error reduces the linearization error in the \ac{mekf} approach improving its performance. The benefit of the \ac{eqf} formulation is, indeed, in the much faster transient, larger beacon of attraction, and consistency rather than in the asymptotic behavior.

\begin{figure}
        \centering
        % \subfloat[Filter energy \label{fig:res_detailed_energy}]{
        \includegraphics[width=\linewidth]{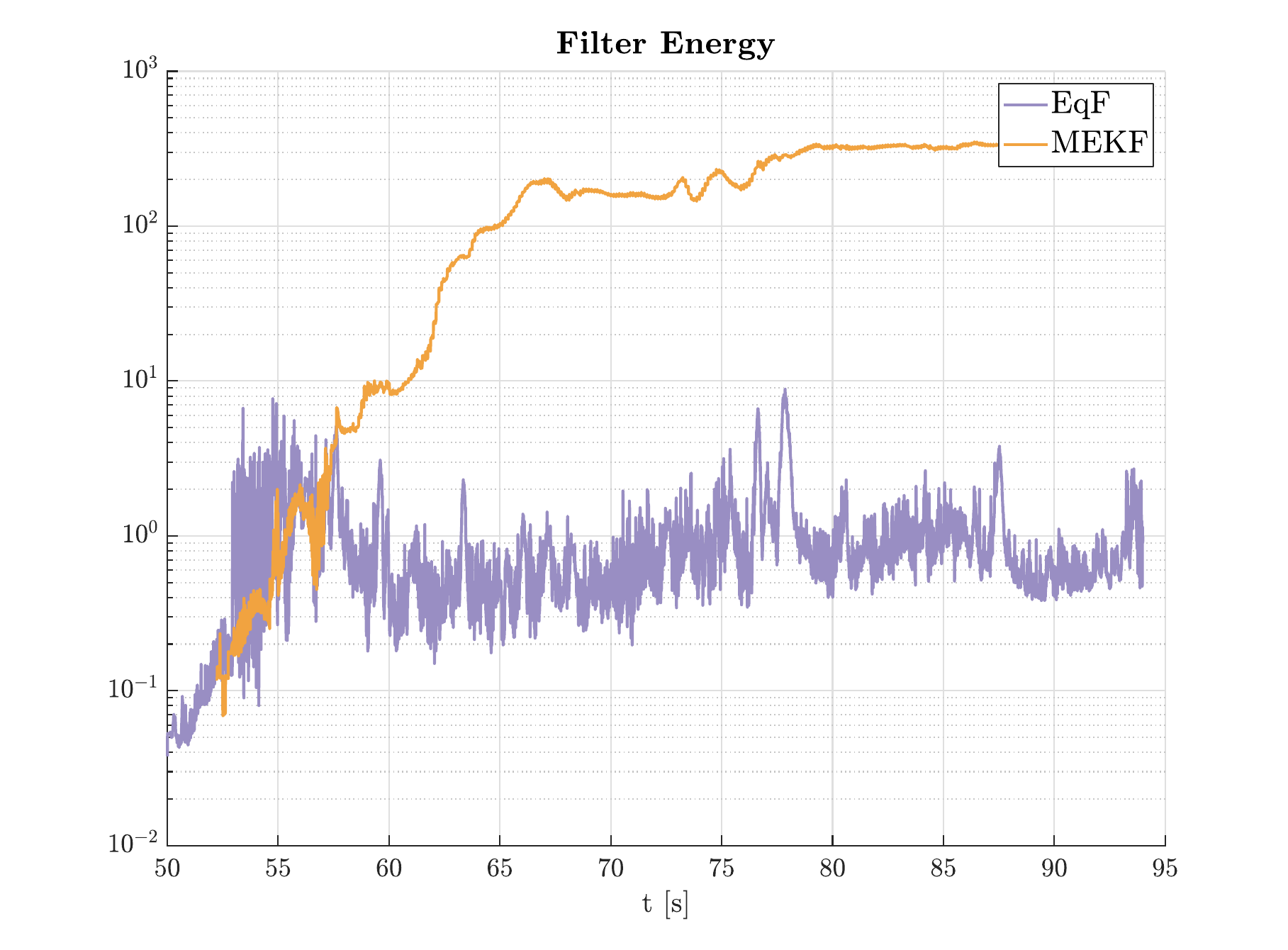}
        % }
        \caption{Filter energy comparison between the \acs{eqf} and \acs{mekf} frameworks on flight \runshort. In this run, the \ac{mekf} is over-confident given its position and calibration state errors. The \ac{eqf} converges to magnitude of $1$ as is expected of consistent filters.}
        %While the \acs{eqf}'s energy is approx. $1$ after the transient phase, the \ac{mekf} is slightly over-confident throughout the flight.}
        \label{fig:res_detailed_energy}
    \end{figure}

    For consistency evaluation, we further present the filter energy (or NEES~\cite{fornasier_vinseval_2021}) for \runshort~in \figt{fig:res_detailed_energy}. For this, we consider the states with available ground-truth, i.e., position, orientation, and both calibration states. For more details on this evaluation and comparisons to many other filter setups we refer the reader to~\cite{fornasier_equivariant_2023}.
    In this evaluation, we clearly see the \ac{eqf}'s filter energy converging to a magnitude of $1$. However, the \ac{mekf} is several order of magnitudes higher, resulting in an over-confident system. We observe this behavior throughout all evaluations of the \emph{INSANE Dataset}.
    %We recognize that the over-confidence of the \ac{mekf} is partly also caused through the wrong initial state and required initial state covariance. In case of correct initialization (e.g., for the \SI{5}{\percent} off ground-truth runs) the energy of the \ac{mekf} resulted in a more consistent filter.
    % While for this evaluation both filters are converge to an energy in the order of magnitude $1$, the \ac{eqf} is on average close to $1$ than the \ac{mekf}, making the latter a slightly over-confident system. In general, throughout the \emph{INSANE dataset} evaluation, we saw the \ac{eqf} perform with a similar confidence, while the \ac{mekf} tends to be over-confident in these evaluations.
    % Clearly, the \ac{eqf} converges to an energy of approx. $1$, which determines a consistent filter. In contrast, the \ac{mekf} formulation is over-confident for the same run.
    %For this implementation, we reckon this might be due to a too large state covariance while not having a large state error. We consider this under-confidence to be a result of the specific implementation, as changing the measurement noise values or initial system covariance did not majorly change the energy result.

    % \renewcommand{\plotrun}{runs_230301}

%------------------------------------------------------------------------------%
\section{Summary}
\label{sec:summary}

With this contribution we provide \iac{eqf} framework for multi-\ac{gnss} setups for \acp{uav}, which takes advantages of a Lie symmetry group for lifting the system dynamics.
We further show that this type of filter formulation yields improved performance compared to the currently most commonly used \acp{mekf}. This is a result of the better system formulation and its inherited properties for convergence, linearization errors, and consistency. When evaluated on multiple runs of the \emph{INSANE Dataset}, the proposed filter is able to correctly estimate the vehicle's pose and all \ac{gnss} extrinsic calibration states throughout all flights.

Overall, this work aims to serve as a basis for future multi-sensor \ac{eqf} frameworks and provide the initial formulation on how a multi-\ac{gnss} setup within a biased-\acl{ins} can be formulated equivariently.

%%%%%%%%%%%%%%%%%%%%%%%%%%%%%%%%%%%%%%%%%%%%%%%%%%%%%%%%%%%%%%%%%%%%%%%%%%%%%%%%
% \section*{Acknowledgment}

%%%%%%%%%%%%%%%%%%%%%%%%%%%%%%%%%%%%%%%%%%%%%%%%%%%%%%%%%%%%%%%%%%%%%%%%%%%%%%%%
\appendices
% \vspace{-1cm}
% \appendix[Differences with IEKF]
\section{Differences with IEKF}
\label{app:iekfdiff}
In this section we highlight the major differences between the proposed \ac{eqf} and the \ac{iekf}. For an in-depth discussion on the topic, we refer the reader to our very recent research~\cite{fornasier_equivariant_2023}.

The \ac{iekf} is a filter design for systems with group affine dynamics on a Lie group~\cite{barrau_invariant_2017}.
The \ac{eqf}, instead, can be seen as a general filter design for systems with symmetries, where the original system is lifted onto a Lie group, the symmetry group, and a filter is designed based on the equivariant error. The \ac{eqf} specializes to the \ac{iekf} for systems with group affine dynamics on a Lie group, when the Lie group is chosen as symmetry group, the origin is chosen to be the identity, and the local coordinates are chosen to be the logarithmic coordinates~\cite[Appendix B]{van_goor_equivariant_2023}.

Previous work has introduced a variant of the \ac{iekf}, coined the ``imperfect-\ac{iekf}''~\cite{barrau_non-linear_2015}, to handle this type of problem.
This paper addresses the problem of filter design for a biased inertial navigation system, as depicted in \equtt{\ref{equ:RGIdt}-\ref{equ:baIdt}}, which does not possess group affine dynamics.
In the imperfect-\ac{iekf}, the IMU biases are treated as elements of a Euclidean space attached to the navigation group, resulting in a Lie group structure of $\SE[2]{3} \times \R^6$. In contrast, our proposed \ac{eqf} exploits the semi-direct product symmetry $\sdgrp$ proposed in~\cite{fornasier_equivariant_2022, fornasier_equivariant_2023}, resulting in a filter with identical error dynamics for attitude, position, and velocity ($\vUp_1$ in \equt{equ:At}), while introducing distinct error dynamics for \ac{imu} biases ($\vUp_2, \vUp_3$ in \equt{equ:At}), thereby improving the linearization error~\cite{fornasier_equivariant_2023}. Additionally, by employing normal coordinates instead of logarithmic coordinates, the proposed \ac{eqf} achieves a third-order linearization error reduction in the output linearization~\cite{van_goor_equivariant_2023}.

% \appendix[INSANE Evaluation]
\section{INSANE Evaluation}
\label{app:insane_evaluation}

\renewcommand{\plotrun}{runs_230301}
\renewcommand{\plotfiletype}{png}

\begin{figure*}[!hb]
        \centering
        \subfloat[Mars 01 \label{fig:mars_01}]{\includegraphics[width=0.24\linewidth]{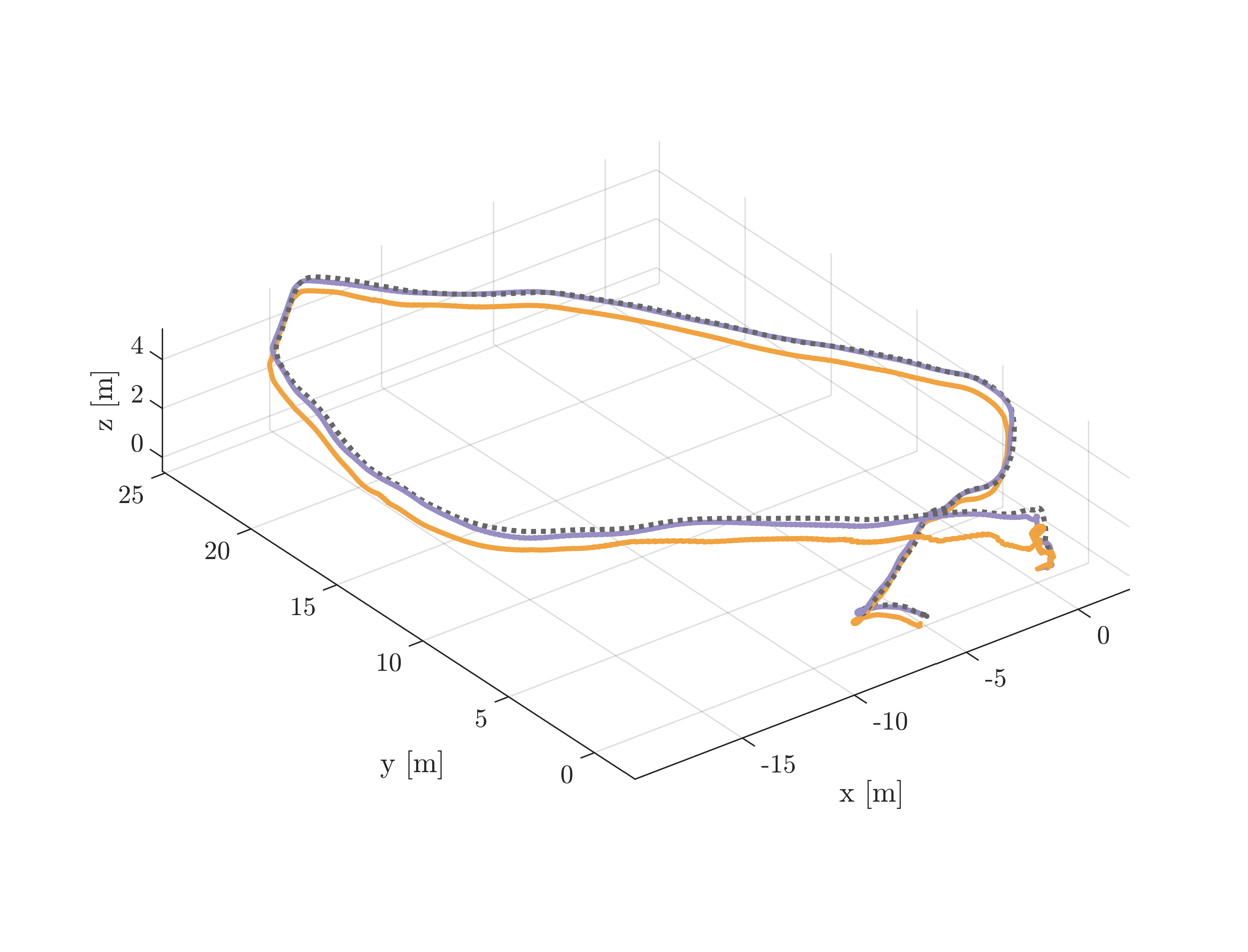}}
        \subfloat[Mars 02 \label{fig:mars_02}]{\includegraphics[width=0.24\linewidth]{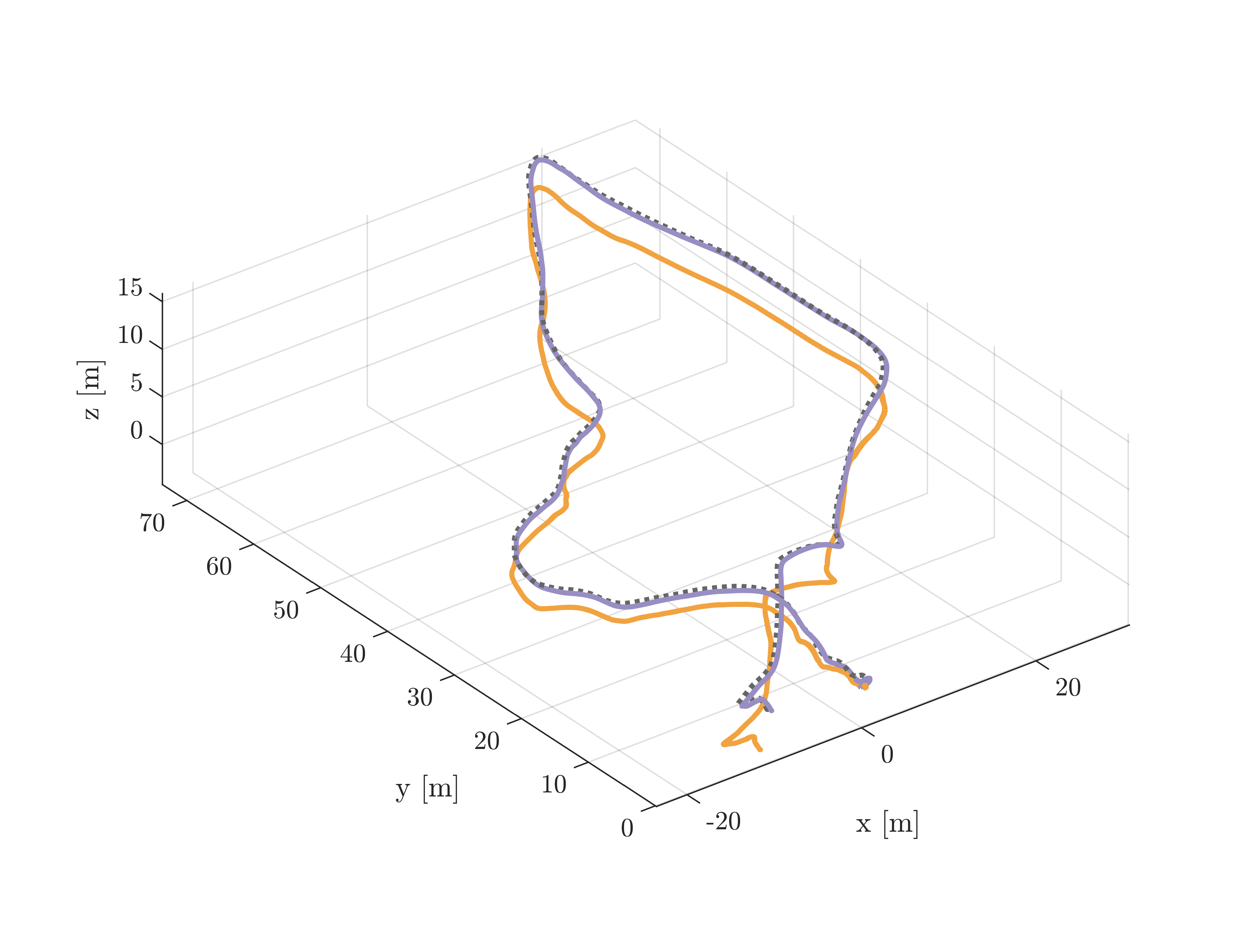}}
        \subfloat[Mars 03 \label{fig:mars_03}]{\includegraphics[width=0.24\linewidth]{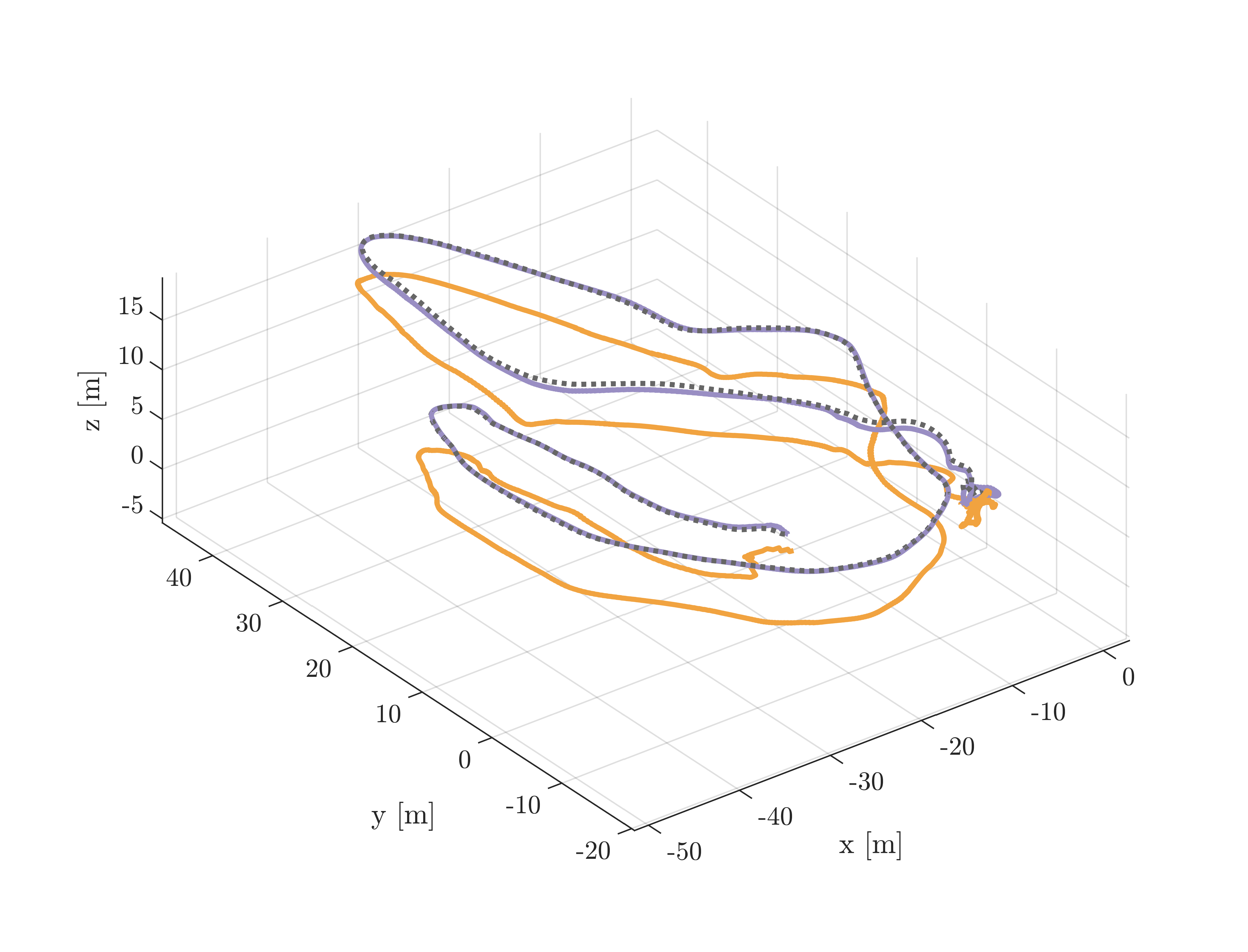}}
        \subfloat[Mars 04 \label{fig:mars_04}]{\includegraphics[width=0.24\linewidth]{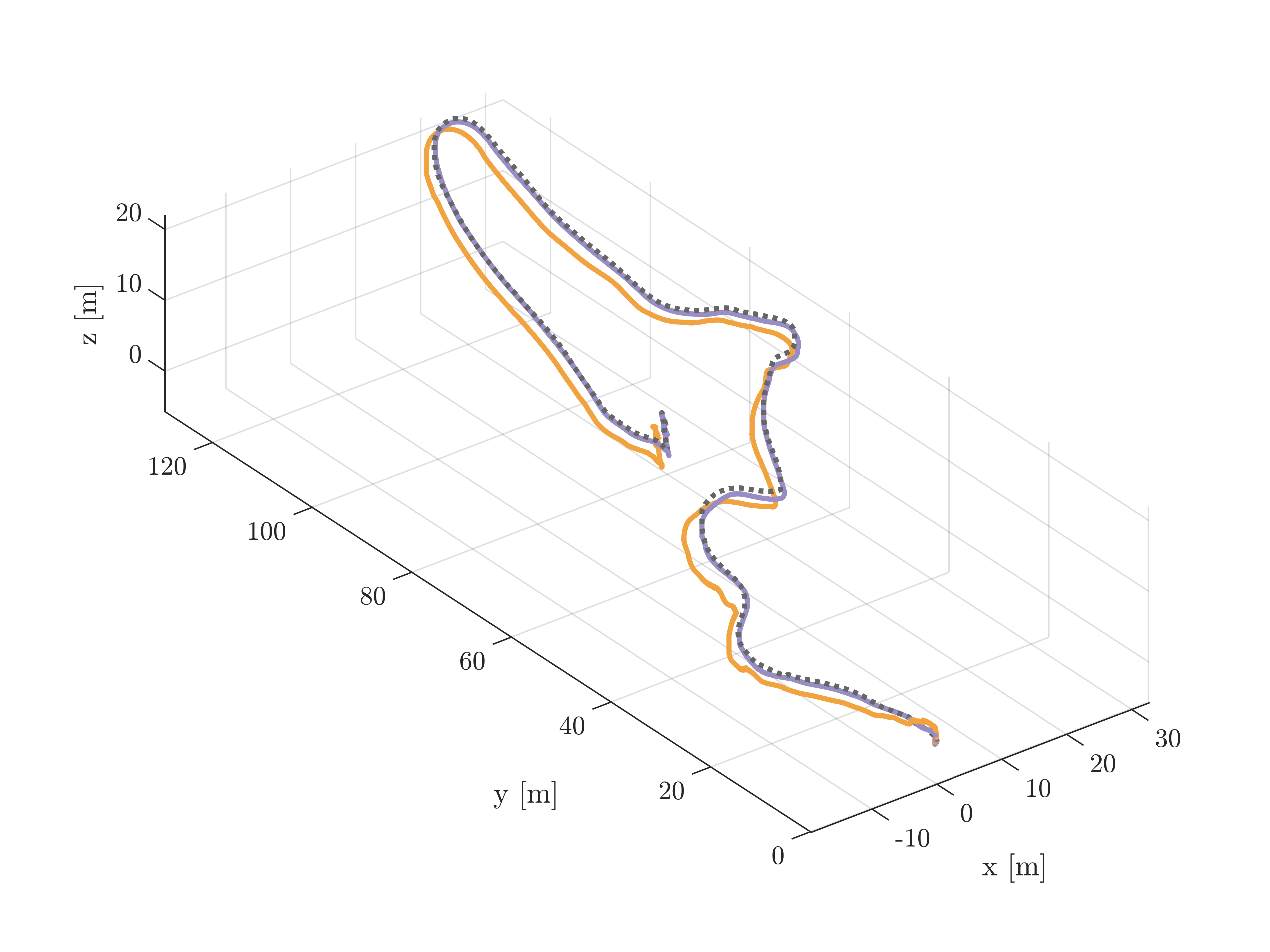}} \\
        \subfloat[Mars 05 \label{fig:mars_05}]{\includegraphics[width=0.24\linewidth]{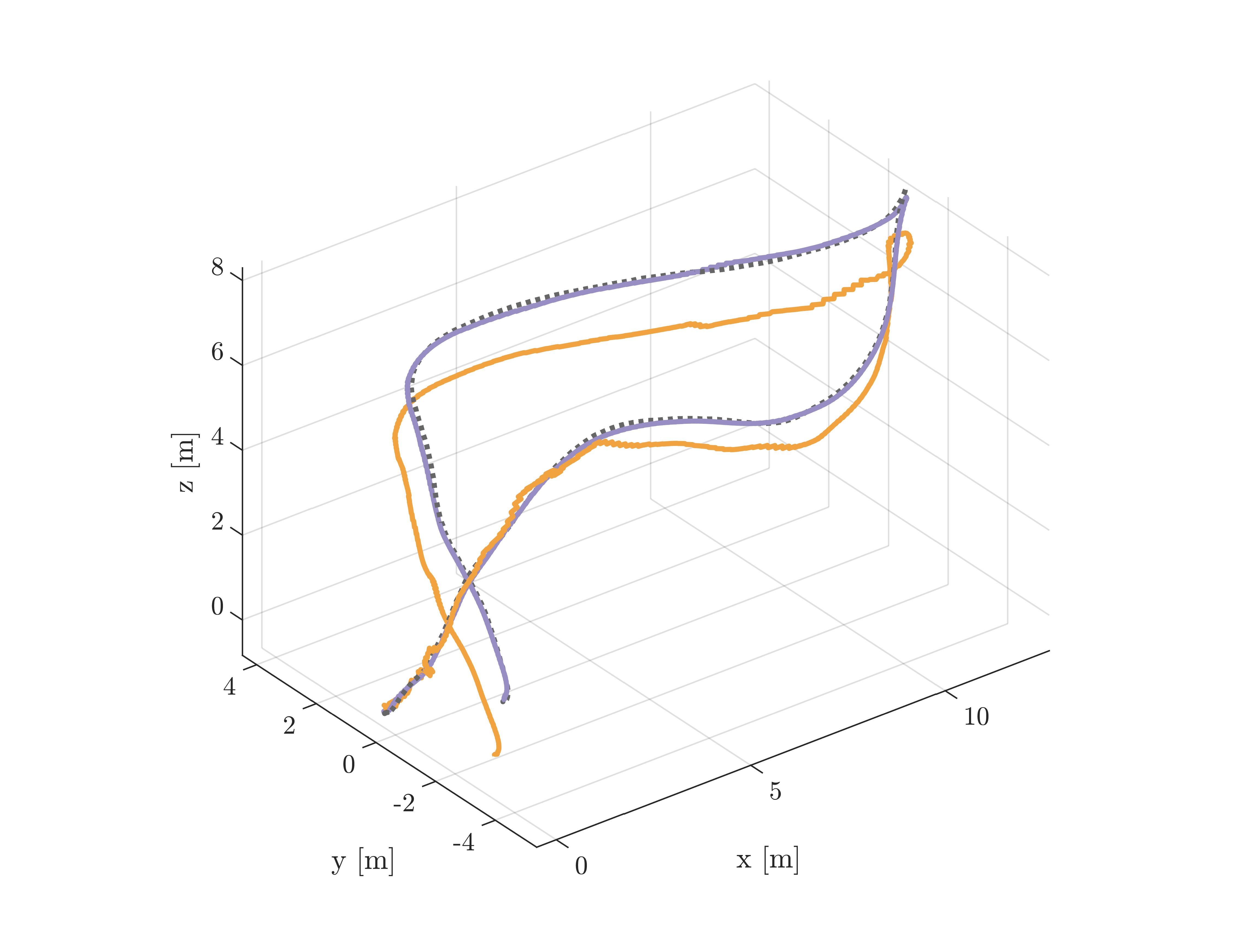}}
        \subfloat[Mars 06 \label{fig:mars_06}]{\includegraphics[width=0.24\linewidth]{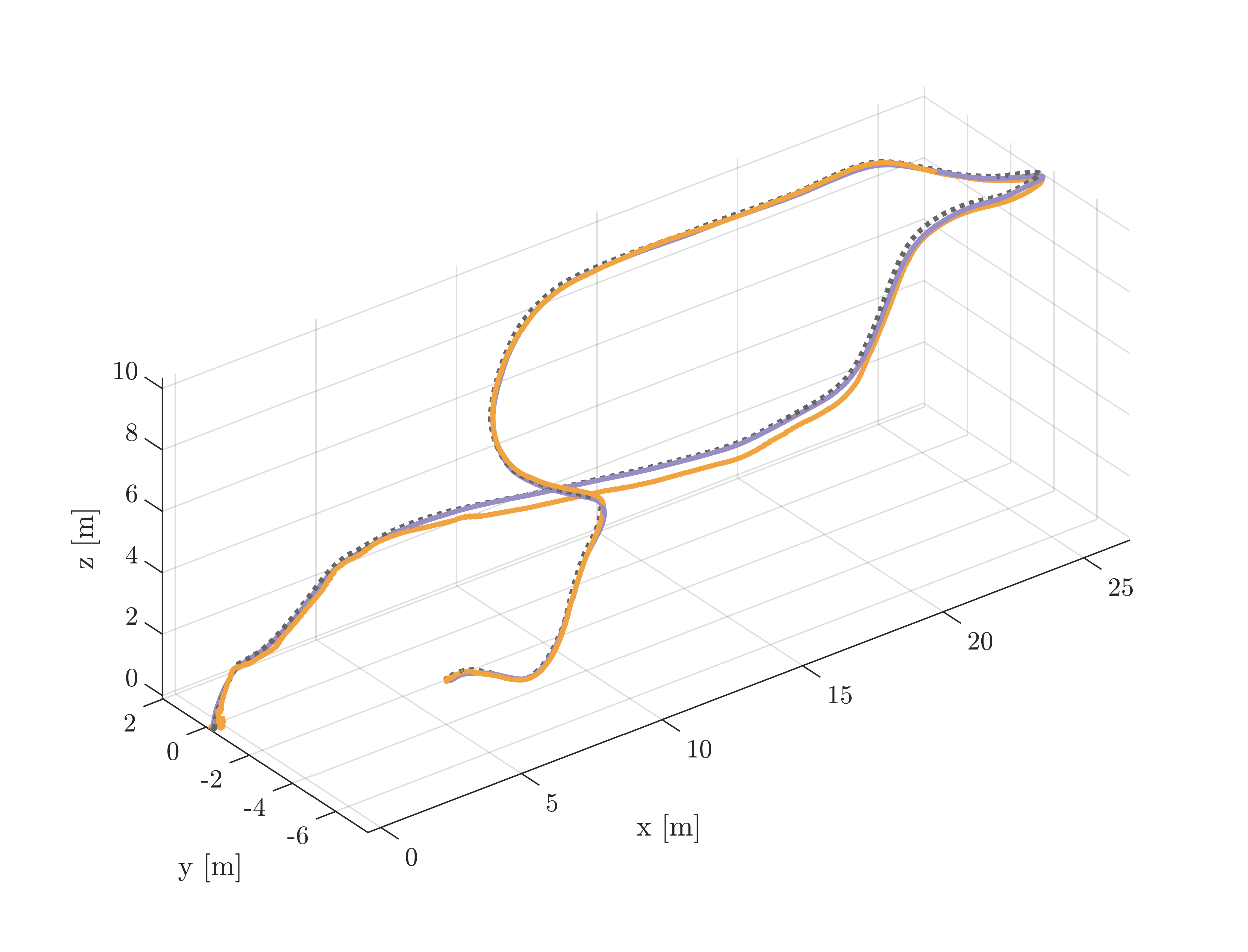}}
        \subfloat[Mars 07 \label{fig:mars_07}]{\includegraphics[width=0.24\linewidth]{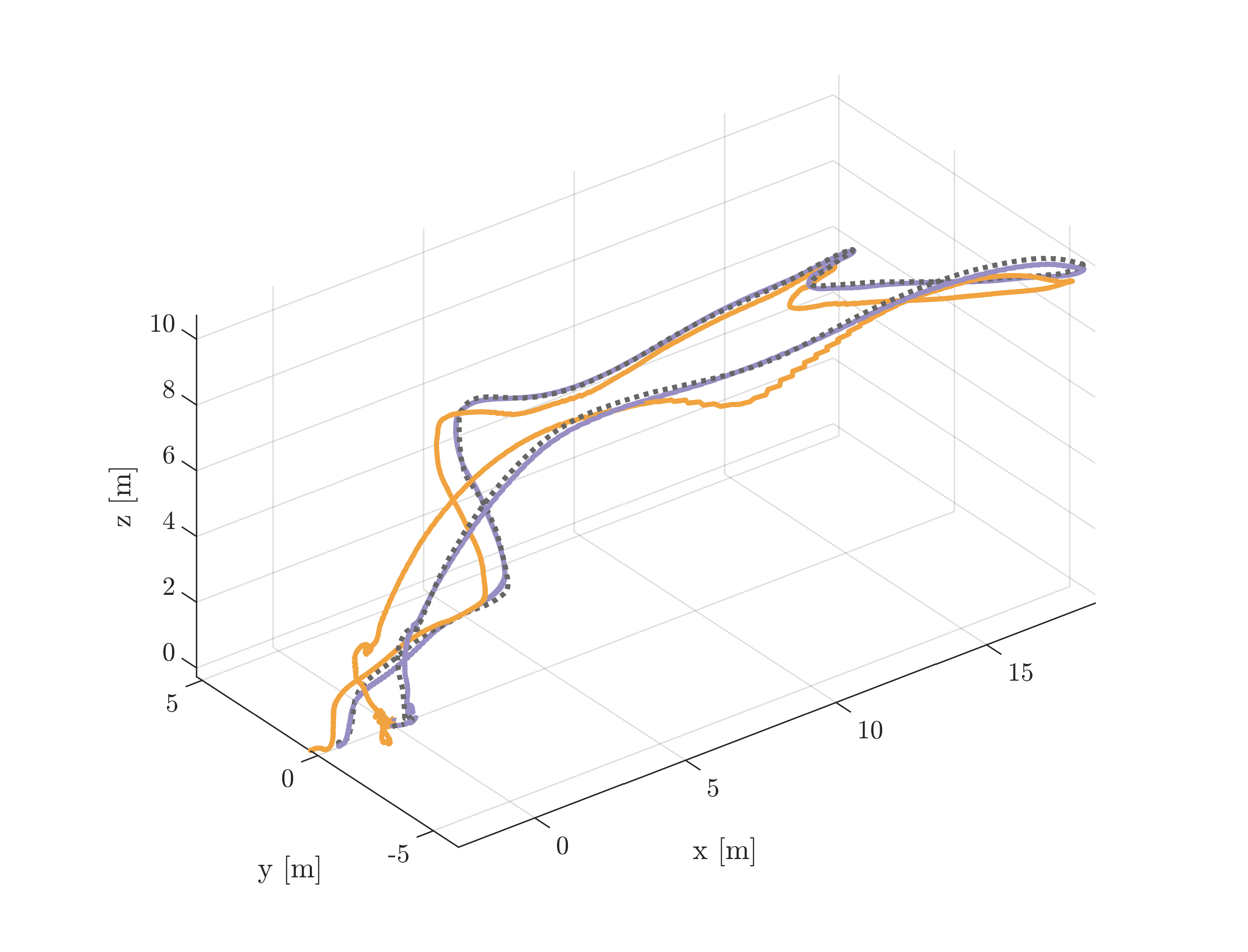}}
        \subfloat[Mars 08 \label{fig:mars_08}]{\includegraphics[width=0.24\linewidth]{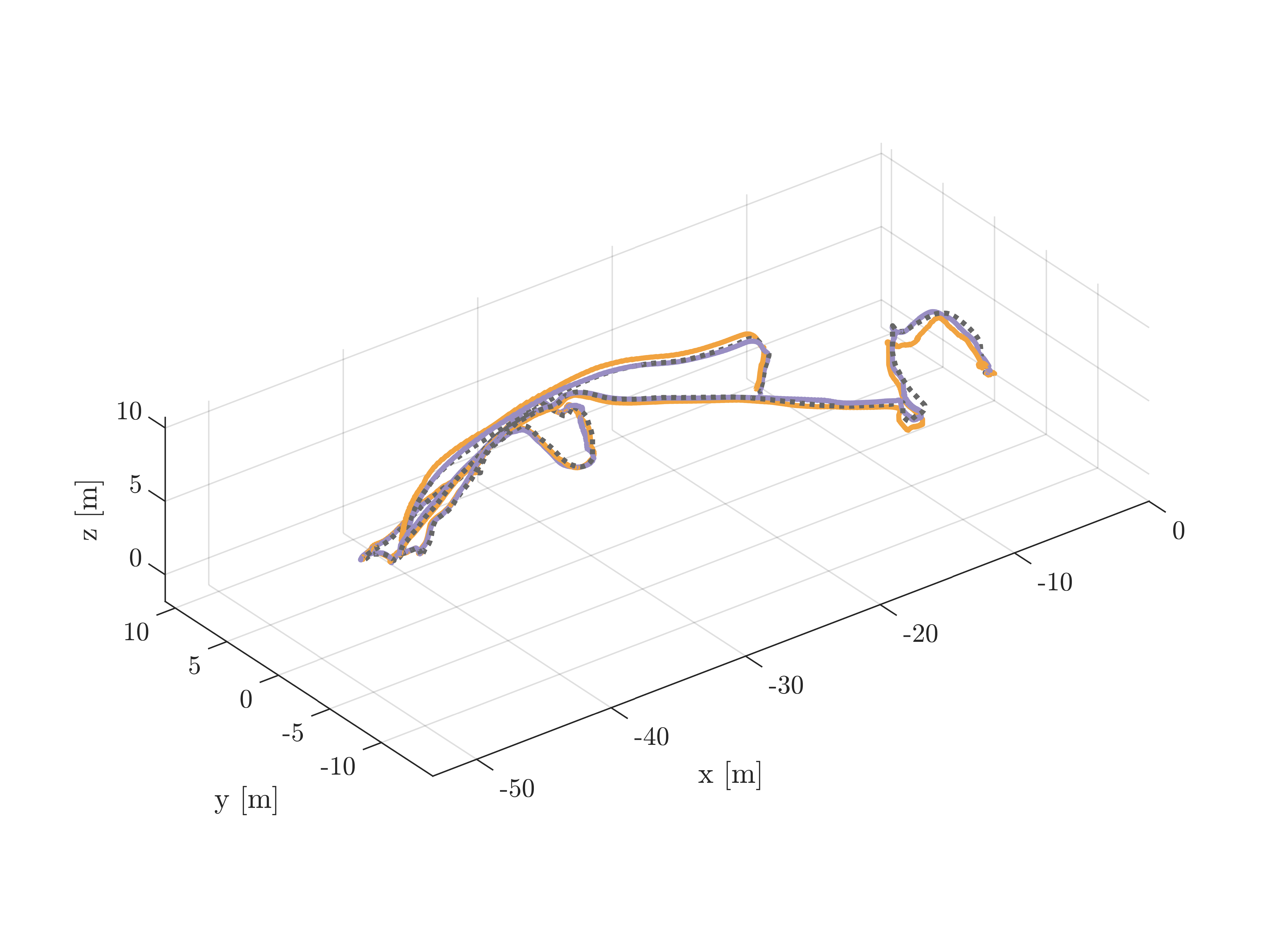}} \\
        \subfloat[Mars 09 \label{fig:mars_09}]{\includegraphics[width=0.24\linewidth]{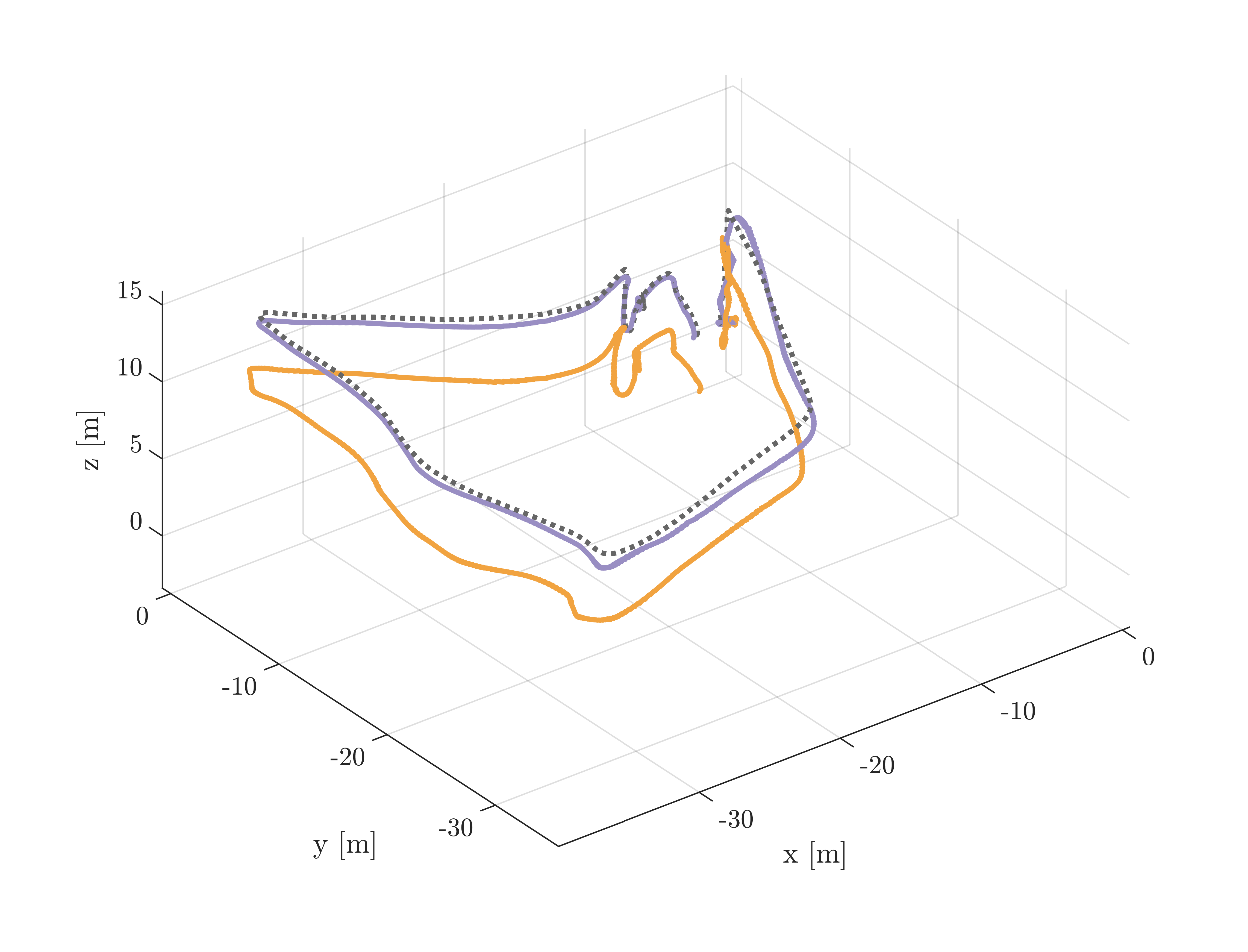}}
        \subfloat[Mars 10 \label{fig:mars_10}]{\includegraphics[width=0.24\linewidth]{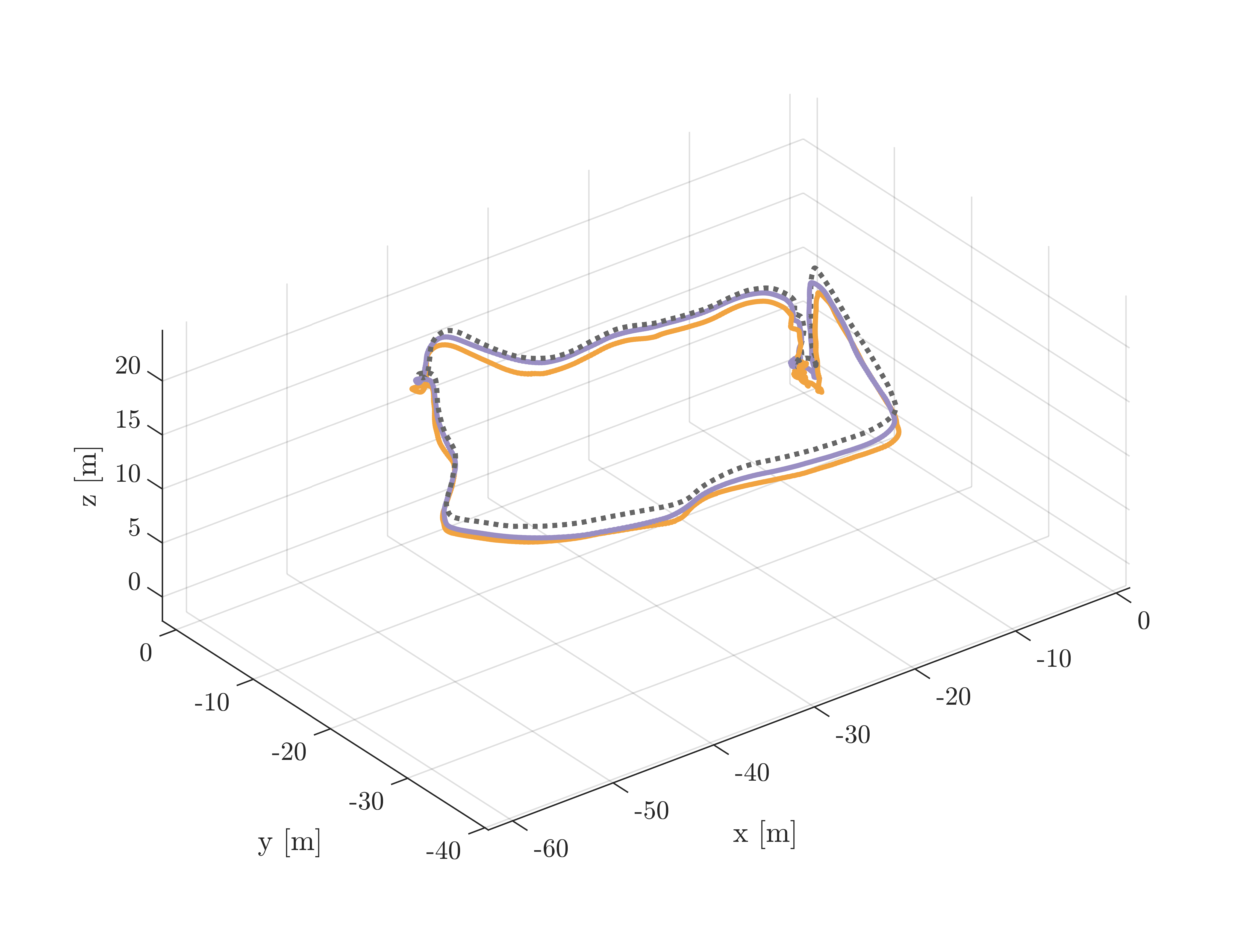}}
        \subfloat[Mars 11 \label{fig:mars_11}]{\includegraphics[width=0.24\linewidth]{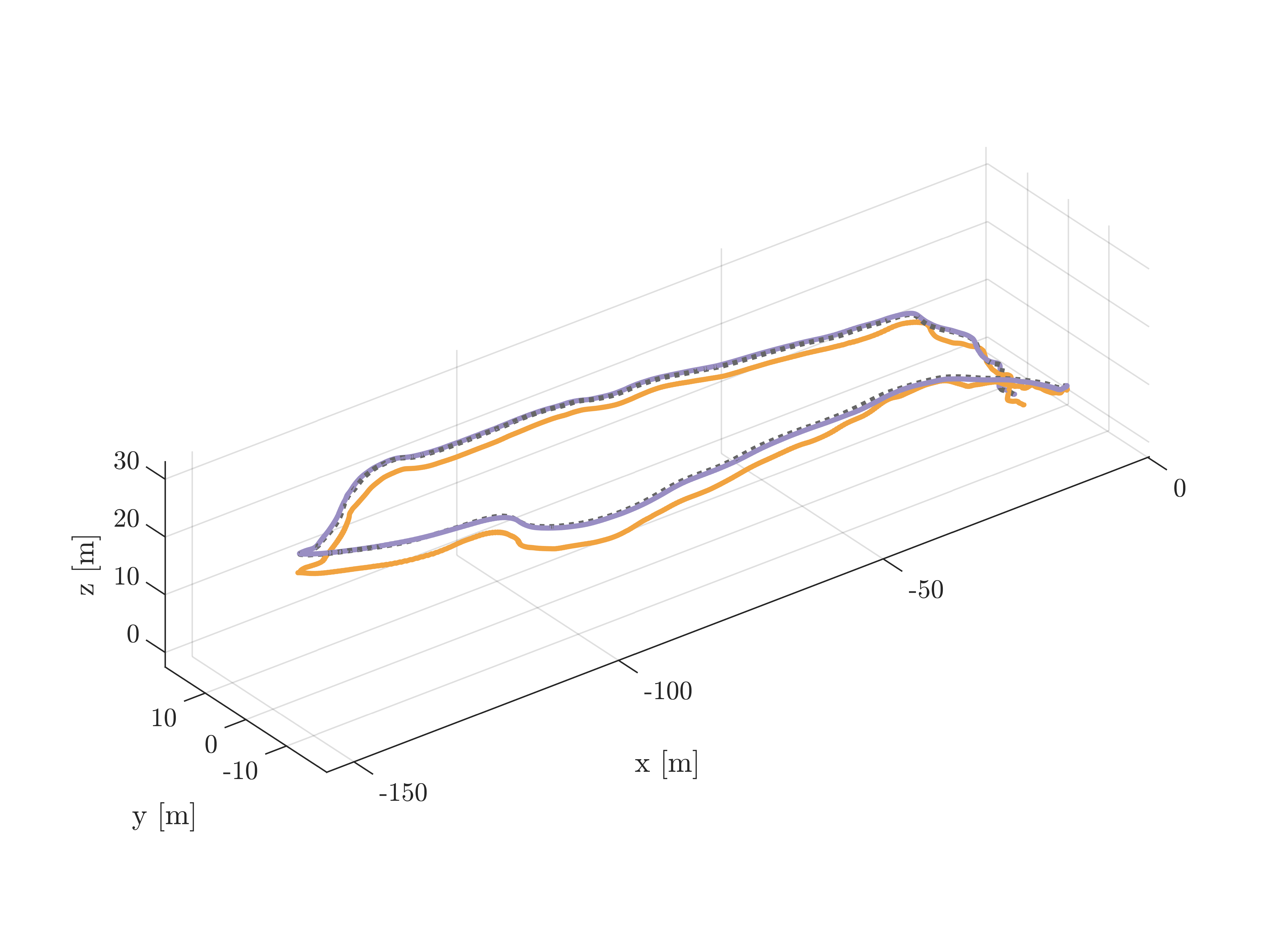}}
        \subfloat[Mars 12 \label{fig:mars_12}]{\includegraphics[width=0.24\linewidth]{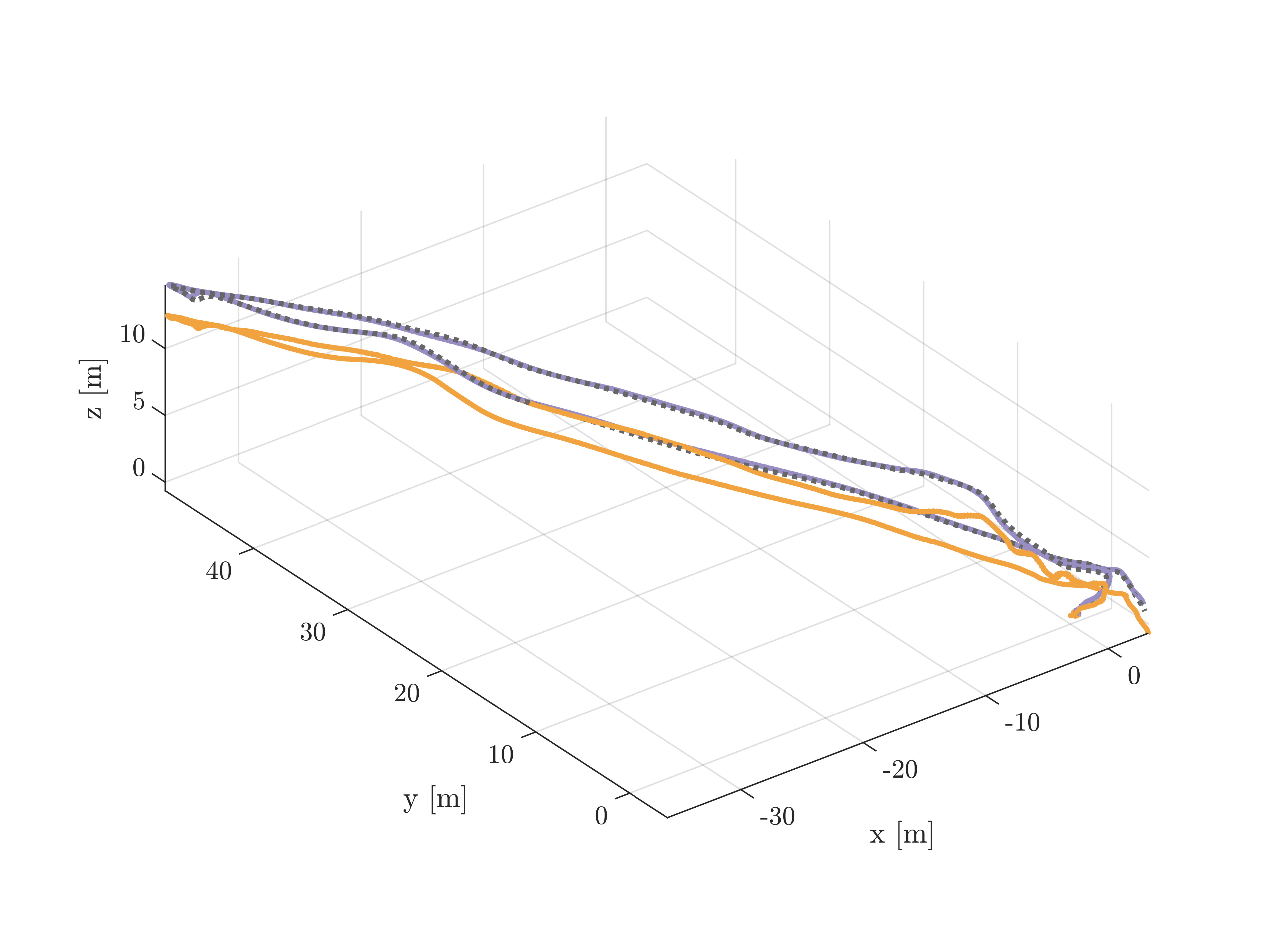}} \\
        \subfloat[Mars 13 \label{fig:mars_13}]{\includegraphics[width=0.24\linewidth]{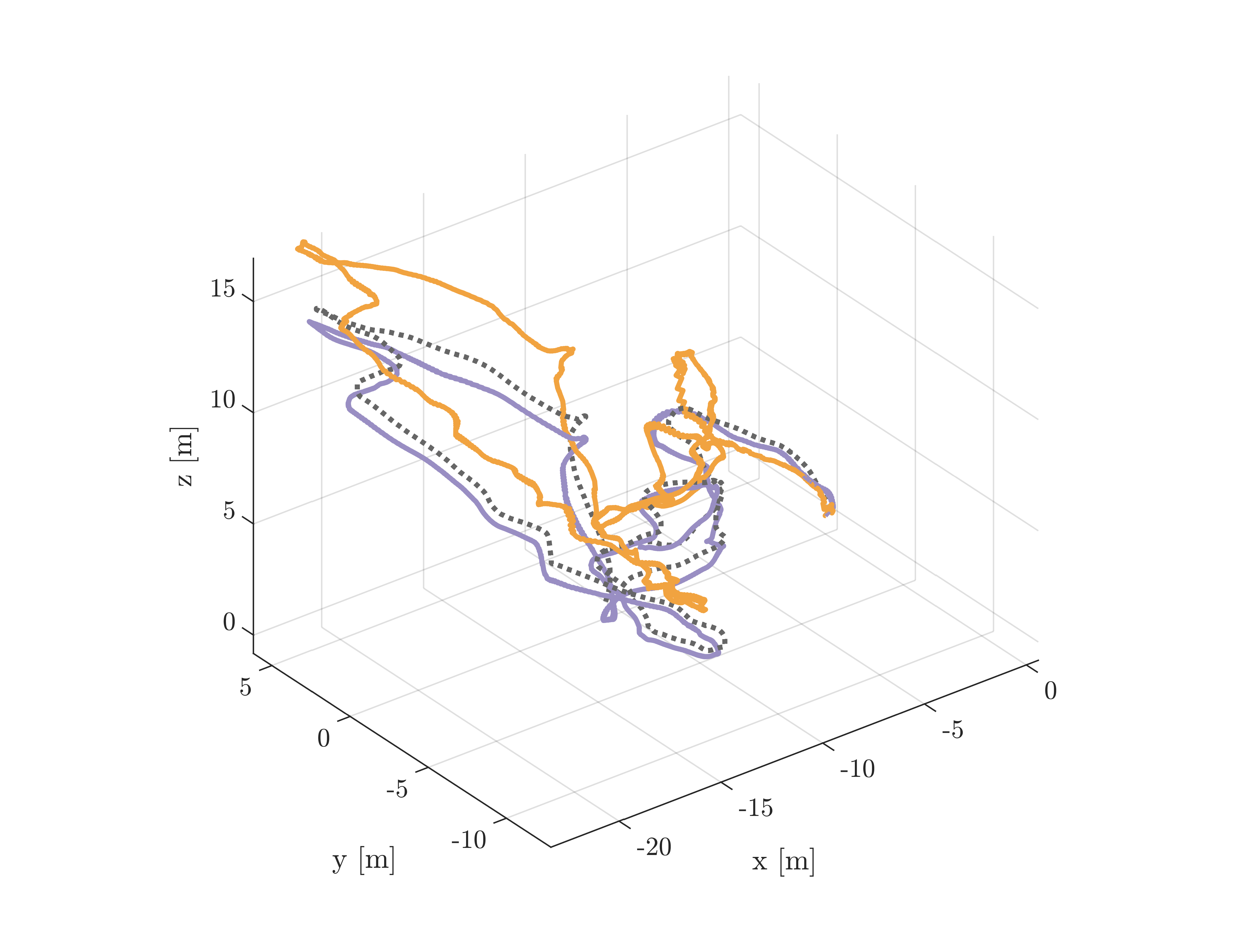}}
        \subfloat[Mars 14 \label{fig:mars_14}]{\includegraphics[width=0.24\linewidth]{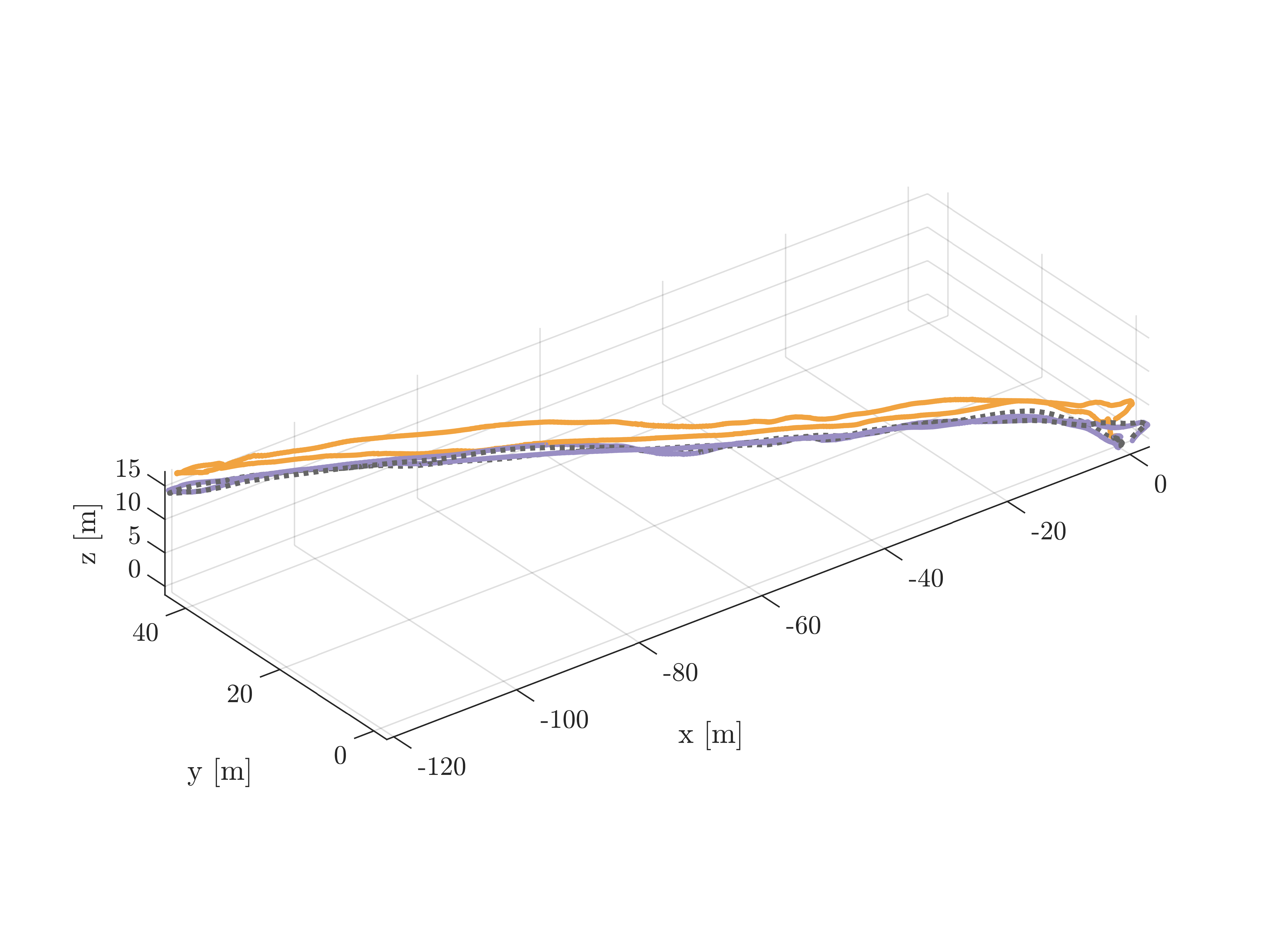}}
        \subfloat[Mars 15 \label{fig:mars_15}]{\includegraphics[width=0.24\linewidth]{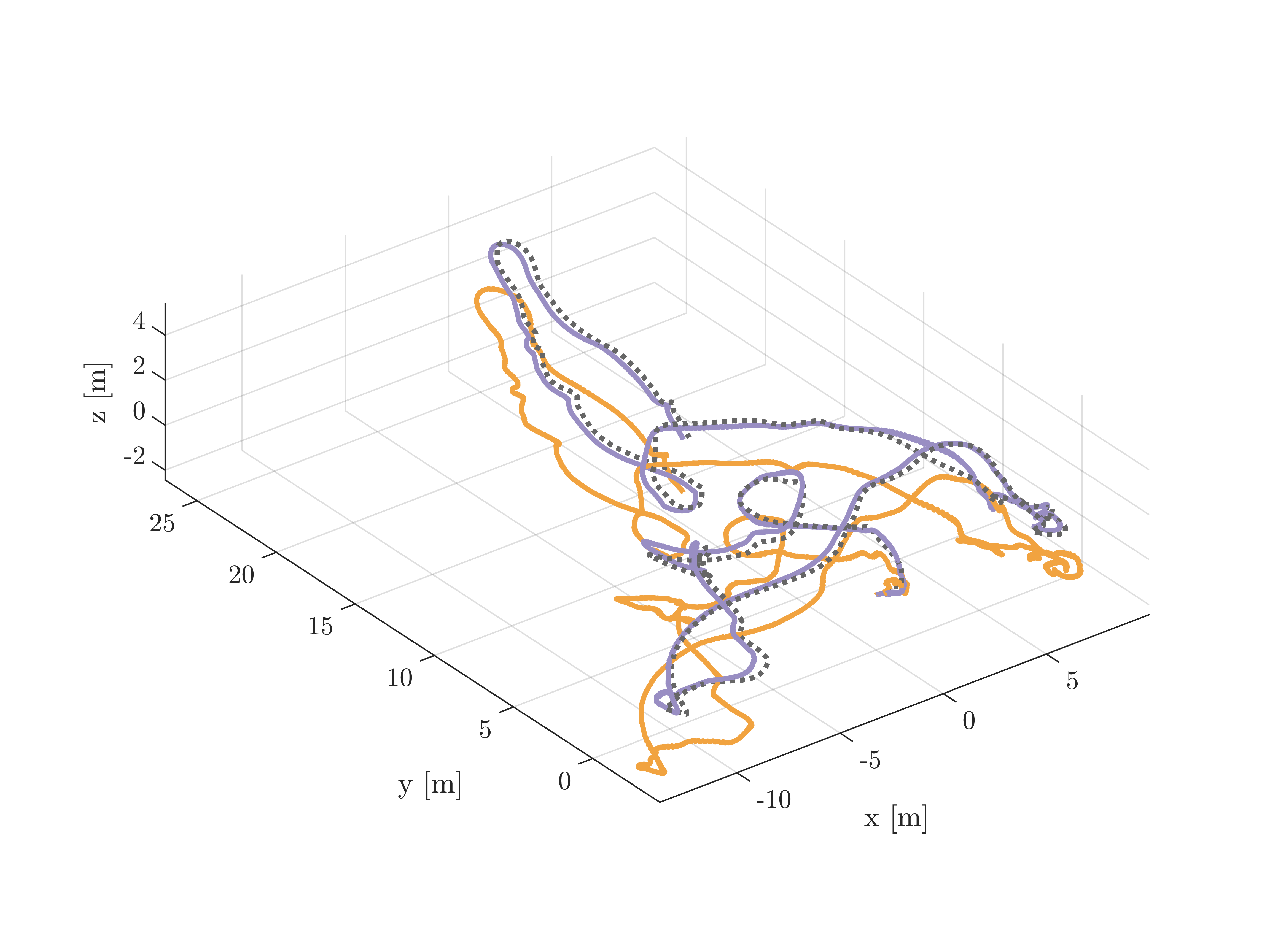}}
        \subfloat[Mars 16 \label{fig:mars_16}]{\includegraphics[width=0.24\linewidth]{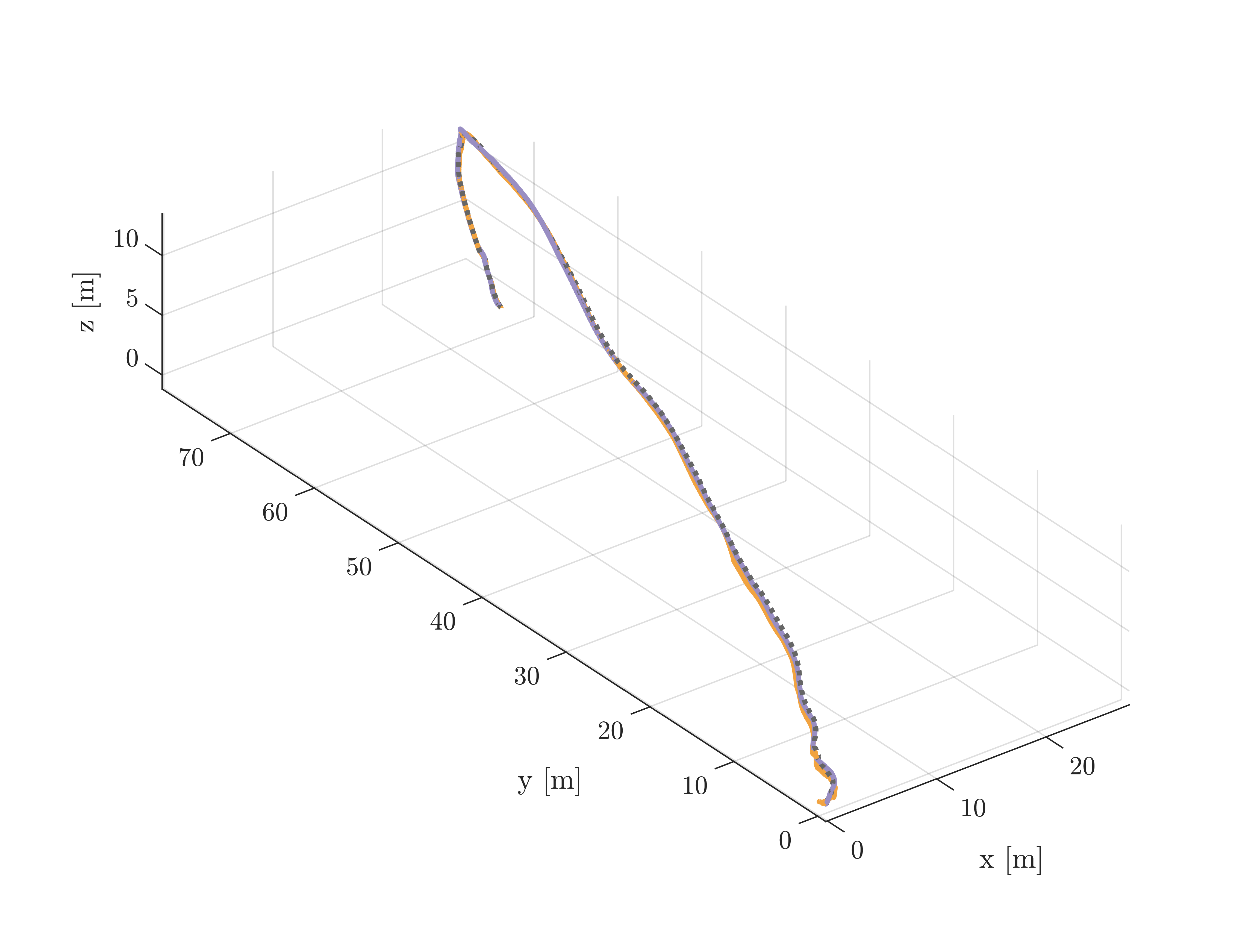}} \\
        %https://www.overleaf.com/project/63e37c9f5cf7effcb5da7021#
        \subfloat[Mars 17 \label{fig:mars_17}]{\includegraphics[width=0.24\linewidth]{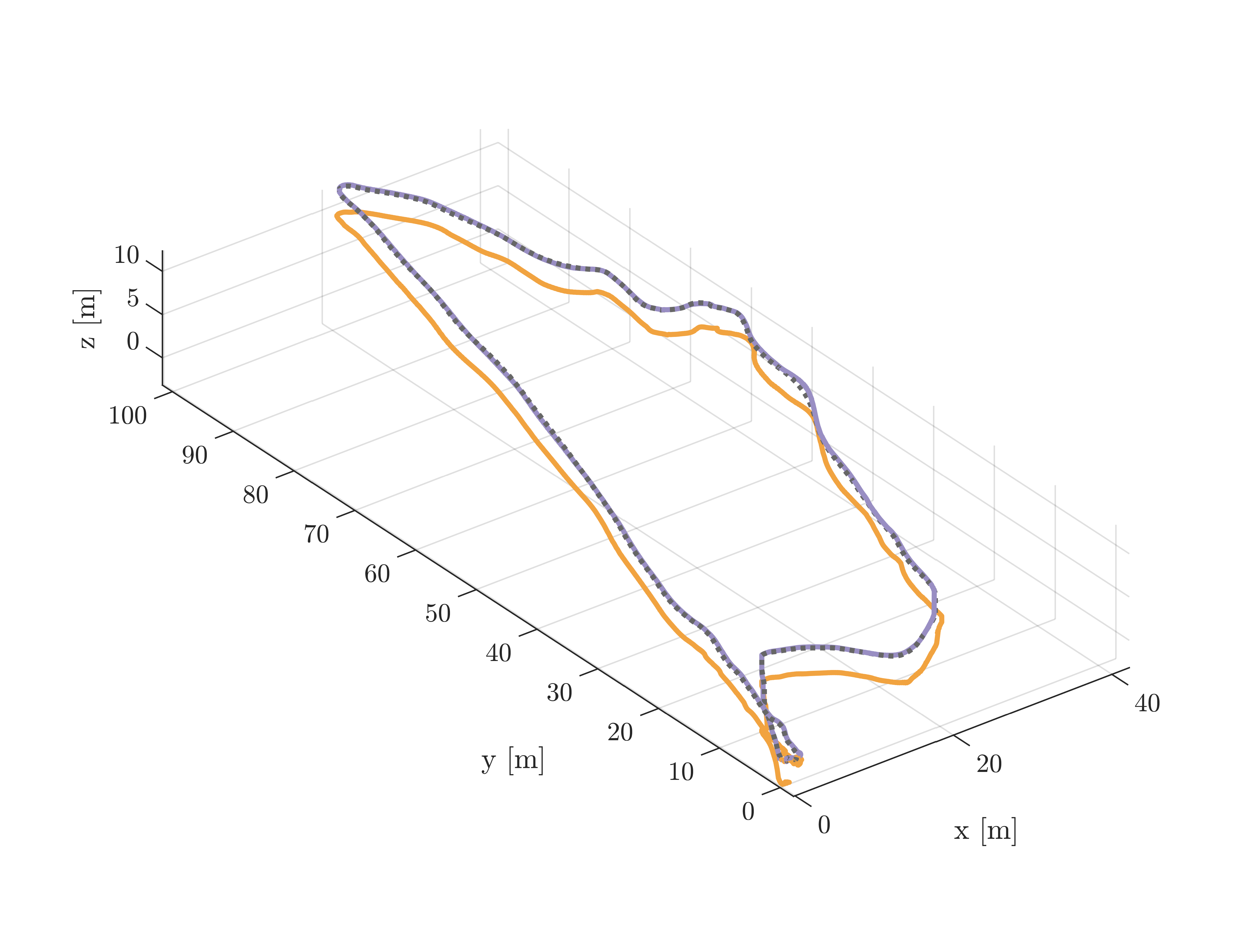}}
        \subfloat[Mars 18 \label{fig:mars_18}]{\includegraphics[width=0.24\linewidth]{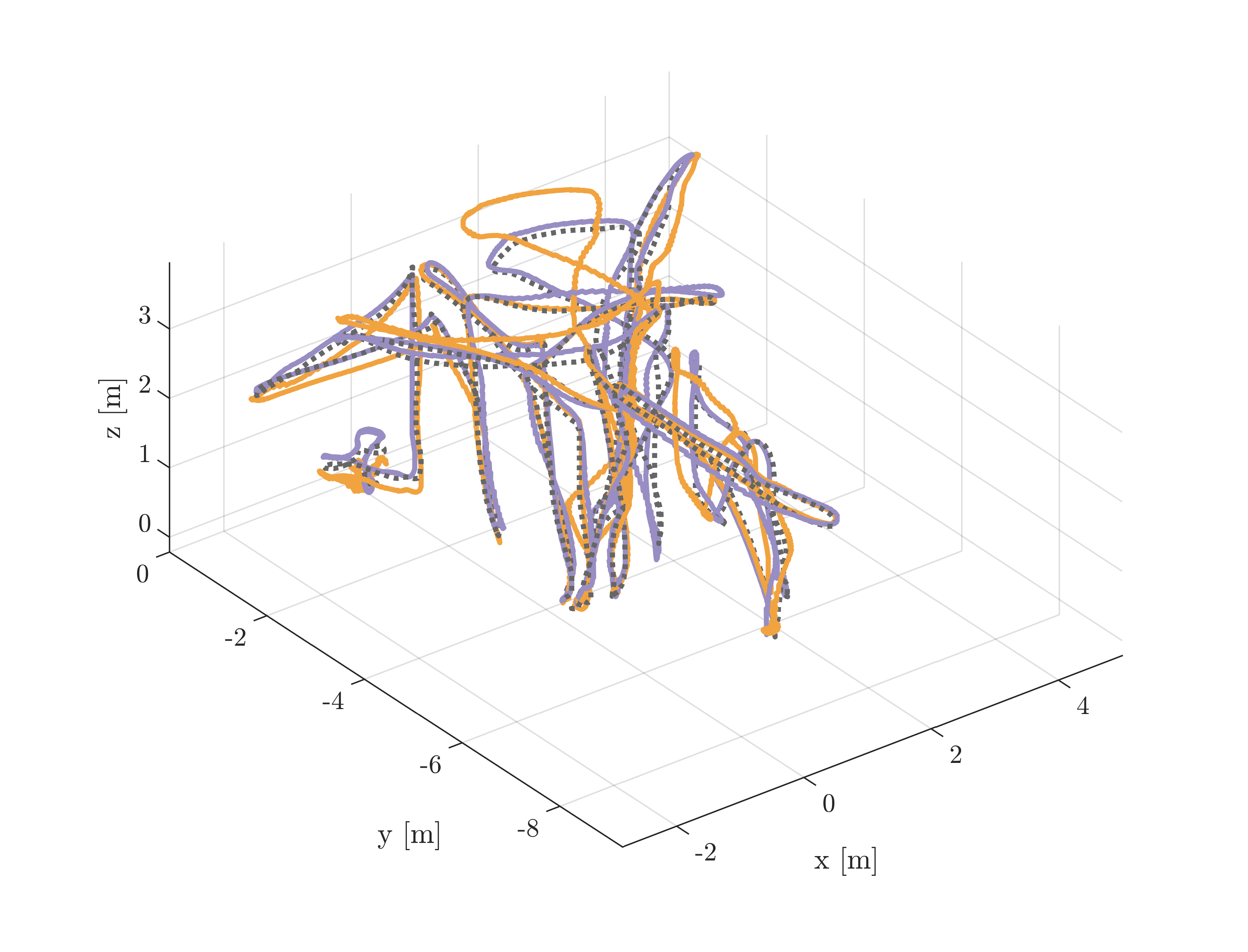}}
        \subfloat[Mars 19 \label{fig:mars_19}]{\includegraphics[width=0.24\linewidth]{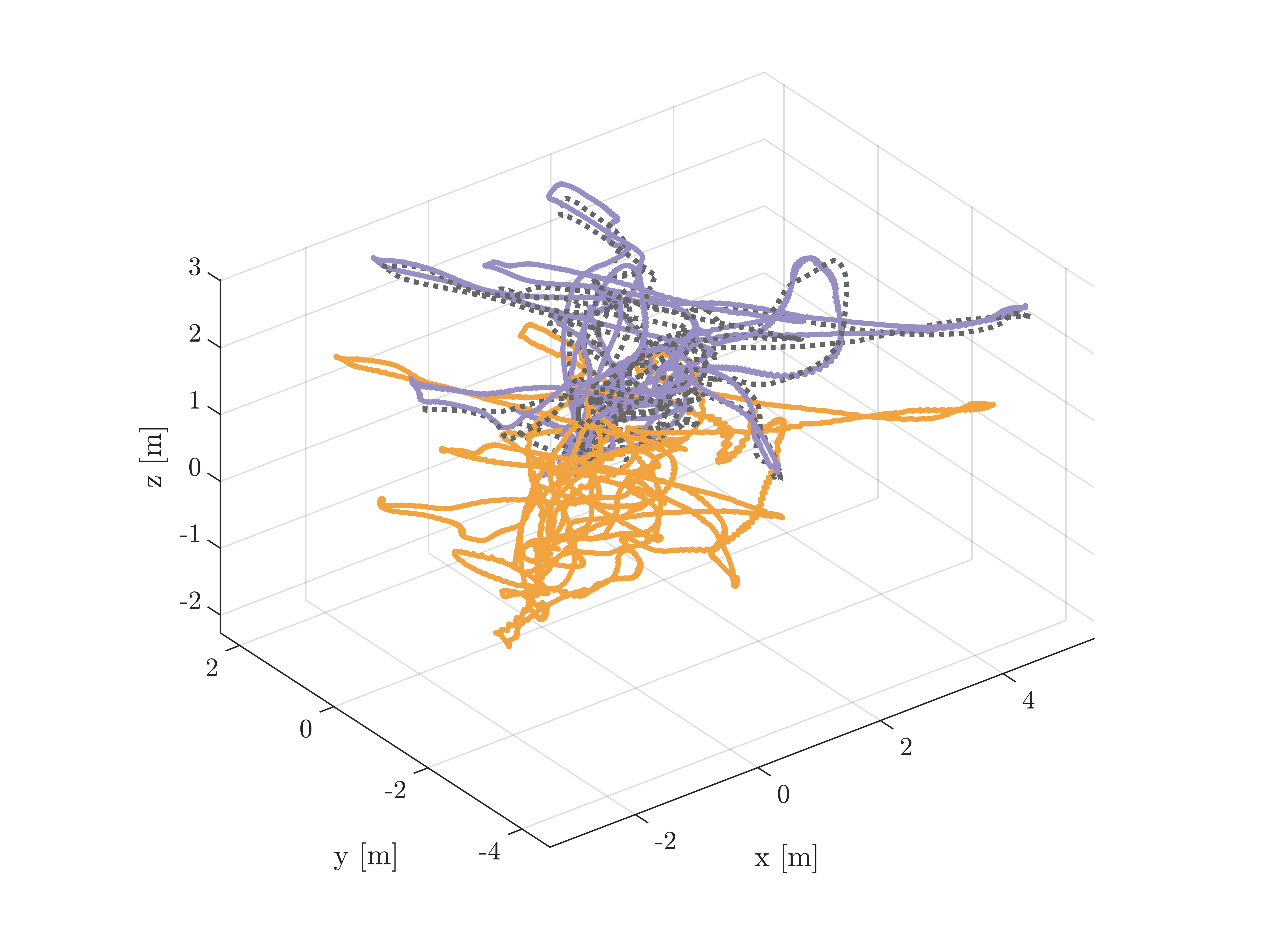}}
        \subfloat[Outdoor 01 \label{fig:outdoor_01}]{\includegraphics[width=0.24\linewidth]{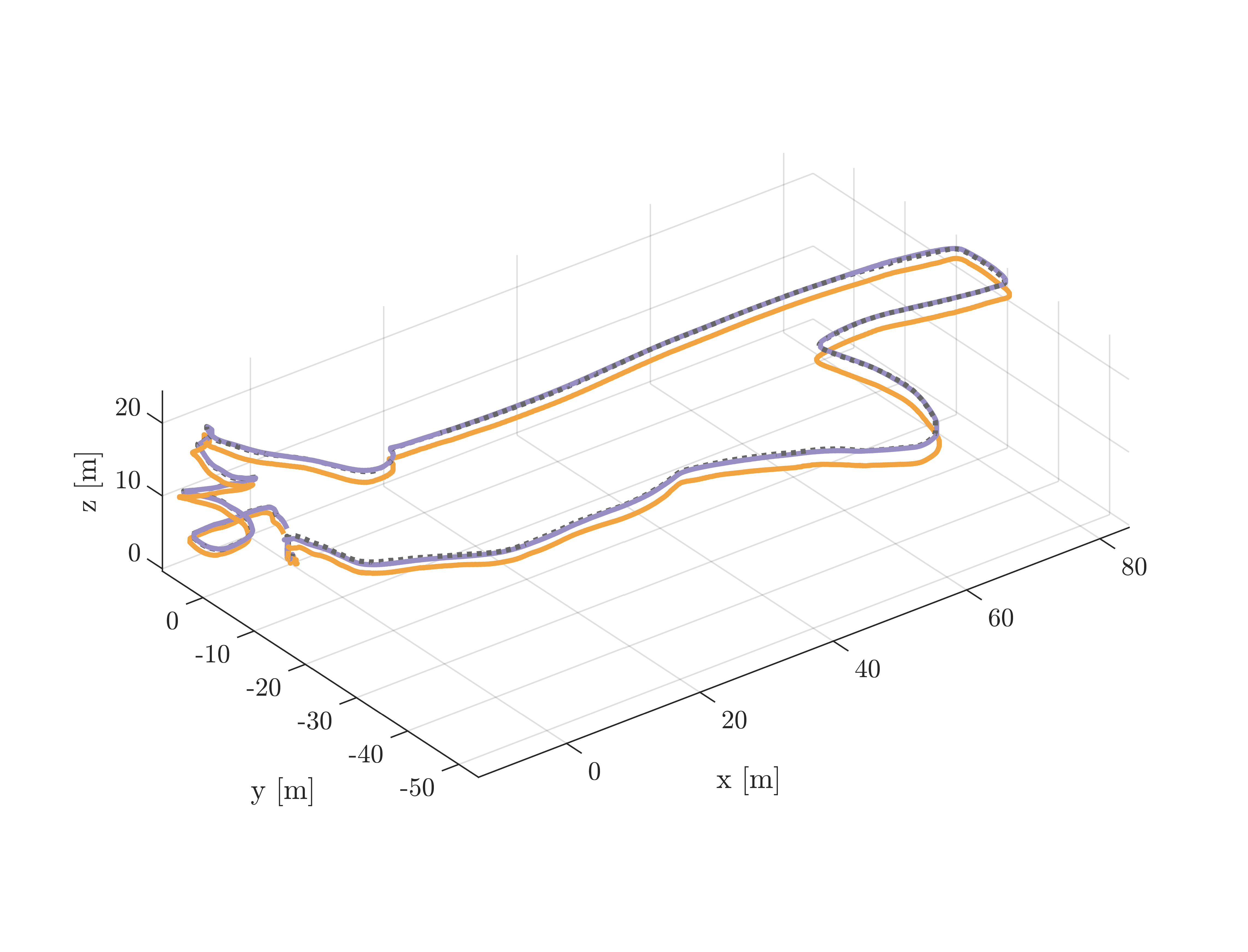}}
        % \missingfigure[figheight=0.5\paperheight]{Images of INSANE runs}
        % \color[rgb]{0.1059, 0.6196, 0.4667}
        % \color[rgb]{0.8510, 0.3725, 0.0078}
        \caption{Evaluation results of the \textbf{\color{PuOr3div3} \acs{eqf} (purple)} and a \textbf{\color{PuOr3div1} \acs{mekf} (yellow)} framework on the \emph{INSANE Dataset}~\cite{brommer_insane_2022}. As displayed in \tabt{tab:rmse} the overall position error is relatively small given the trajectory lengths. Nonetheless, in most runs the \acs{eqf} outperforms a \acs{mekf} implementation.}%, with the exception of run M10 (\figt{fig:mars_10}).}
        \label{fig:results_insane}
    \end{figure*}

For completeness we provide the evaluation of the proposed  \ac{eqf} framework on all 19 ``\textit{Mars}'' and the ``\textit{Outdoor}'' flights of the \emph{INSANE Dataset}~\cite{brommer_insane_2022}. The evaluation of each individual flight is shown in \figt{fig:results_insane}. With this evaluation, we can successfully demonstrate that \iac{eqf} framework performs successful state estimation in real-world environments.

%%%%%%%%%%%%%%%%%%%%%%%%%%%%%%%%%%%%%%%%%%%%%%%%%%%%%%%%%%%%%%%%%%%%%%%%%%%%%%%%
% \clearpage
% \newpage
\bibliographystyle{styles/ieee/IEEEtran}
% \bibliography{refs/scm_zotero_papers,refs/eqf,refs/temporary_bibs}
\diff{\bibliography{refs/scm_zotero_papers}}
% \balancePageNum{7}

%% FINAL PAGE

\end{document}